\documentclass[10pt,twocolumn,letterpaper]{IEEEtran}

\usepackage{iccv}
\usepackage{times}
\usepackage{epsfig}
\usepackage{graphicx}
\usepackage{amsmath}
\usepackage{amssymb}
\usepackage{ulem}

\usepackage[pagebackref=true,breaklinks=true,letterpaper=true,colorlinks,bookmarks=false]{hyperref}
\usepackage{multirow}

\iccvfinalcopy % *** Uncomment this line for the final submission
\vspace{-0.1in}
\def\iccvPaperID{10211} % *** Enter the ICCV Paper ID here

\ificcvfinal\fi

\newcommand\dataset[0]{LAION-L\textsuperscript{2}I\xspace}
\newcommand\aemodel[0]{CaT\textsuperscript{2}I-AE\xspace}
\usepackage{xcolor}

\begin{document}
\twocolumn[{%
\renewcommand\twocolumn[1][]{#1}%
%%%%%%%%% TITLE
\title{Text2Layer: Layered Image Generation using Latent Diffusion Model}
\vspace{-0.1in}
% \author{Xinyang Zhang \qquad Wentian Zhao \qquad Jeff Chien \qquad Xin Lu \\
% Adobe System, Inc\\
% % Institution1 address\\
% {\tt\small xinyangz,wezhao,jchien,xinl@adobe.com}
% % For a paper whose authors are all at the same institution,
% % omit the following lines up until the closing ``}''.
% % Additional authors and addresses can be added with ``\and'',
% % just like the second author.
% % To save space, use either the email address or home page, not both
% % \and
% % Second Author\\
% % Institution2\\
% % First line of institution2 address\\
% % {\tt\small secondauthor@i2.org}
% }
\author{
    \IEEEauthorblockA{Xinyang Zhang, Wentian Zhao, Xin Lu, and Jeff Chien} \\
\IEEEauthorblockA{Adobe Inc.\\ \texttt{\{xinyangz, wezhao, xinl, jchien\}@adobe.com}}
}
% \author{Xinyang Zhang \qquad Wentian Zhao \qquad Jeff Chien \qquad Xin Lu \\
% Adobe System, Inc\\
% % Institution1 address\\
% {\tt\small xinyangz,wezhao,jchien,xinl@adobe.com}
% % For a paper whose authors are all at the same institution,
% % omit the following lines up until the closing ``}''.
% % Additional authors and addresses can be added with ``\and'',
% % just like the second author.
% % To save space, use either the email address or home page, not both
% % \and
% % Second Author\\
% % Institution2\\
% % First line of institution2 address\\
% % {\tt\small secondauthor@i2.org}
% }
\vspace{-0.1in}
\maketitle
\begin{center}
    \centering
    \vspace{-0.3in}
    \includegraphics[width=0.98\textwidth]{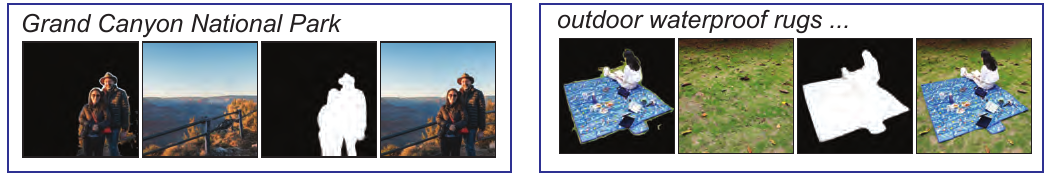}
    % \includegraphics[width=0.98\textwidth]{figures/dataset/dataset_examples_ss.pdf}

    %\vspace{-0.1in}
    \captionof{figure}{Examples of two-layer images. Prompts are displayed on the top of images. Each example includes foreground~(fg), background~(bg), and mask component to compose a two-layer image. From left to right of each example: fg, bg, mask, and composed image.}
    \label{fig:dataset_examples}
    %\vspace{-0.1in}
\end{center}
}]
% Remove page # from the first page of camera-ready.
\ificcvfinal\thispagestyle{empty}\fi

%%%%%%%%% ABSTRACT
\begin{abstract}
\vspace{-0.1in}
Layer compositing is one of the most popular image editing workflows among both amateurs and professionals. %Conventional editing tools allow users to either rely on segmentation/matte tools or manually create layer masks, and then manually replace background with blending modes and harmonization. 
%Diffusion based editing methods automatically edit the desired region by text-guided generation, which usually adapts both the foreground and background without an explicit mask. However, keeping the undesired region untouched is challenging for diffusion based editing due to the sequential application of denoising autoencoders. 
Motivated by the success of diffusion models, we explore layer compositing from a layered image generation perspective. Instead of generating an image, we propose to generate background, foreground, layer mask, and the composed image simultaneously. To achieve layered image generation, we train an autoencoder that is able to reconstruct layered images and train diffusion models on the latent representation. One benefit of the proposed problem is to enable better compositing workflows in addition to the high-quality image output. Another benefit is producing higher-quality layer masks compared to masks produced by a separate step of image segmentation. Experimental results show that the proposed method is able to generate high-quality layered images and initiates a benchmark for future work.
\end{abstract}

%%%%%%%%% BODY TEXT

\vspace{-0.2in}
\section{Introduction}
%Conventional editing tools allow users to either rely on segmentation/matte tools or manually create layer masks, and then manually replace background with blending modes and harmonization. 
%Diffusion based editing methods automatically edit the desired region by text-guided generation, which usually adapts both the foreground and background without an explicit mask. However, keeping the undesired region untouched is challenging for diffusion based editing due to the sequential application of denoising autoencoders. 

Segmenting an image into layers is essential for image editing applications, such as background replacement and background blur. For many years, the development, therefore, has been focused on automatic approaches for various of image segmentation and matting problems~\cite{tpami:2007:closedform,cvpr:2017:deepmat,miccai:2015:unet,cvpr:2017:mask-rcnn}. Recent development rethinks image editing problems from a generative perspective~\cite{nips:2021:editgan,cvpr:2021:lang-guided-edit,arxiv:2022:instructpix2pix,arxiv:2022:dreambooth} and aims to predict the expected editing region implicitly. For example, Text2live~\cite{eccv:2022:text2live} aims to generate desired edits by generating a new image layer guided by text to realize non-disruptive editing. T2ONet~\cite{cvpr:2021:lang-guided-edit} proposes to directly generate image editing results guided by text through generative models. These approaches are limited by the quality of the implicitly predicted editing region and limited by the generated image quality.

Recent efforts in large-scale text2image diffusion models, such as DALL·E 2~\cite{arxiv:2022:dalle2}, Imagen~\cite{arxiv:2022:imagen}, and Stable Diffusion~\cite{cvpr:2022:ldm} significantly improve the image generation quality and produce high-resolution photo-realistic images. Diffusion-based automatic image editing approach is emerging~\cite{cvpr:2022:blended-diffusion,iclr:2022:sdedit,arxiv:2022:dreambooth,cvpr:2022:diffusionclip,arxiv:2022:imagic,arxiv:2022:prompt-to-prompt,arxiv:2022:instructpix2pix,iclr:2023:textual-inversion,arxiv:2022:sine}. For example, Prompt-to-Prompt~\cite{arxiv:2022:prompt-to-prompt} modulates the cross-attention maps in the inference steps to manipulate images. Soon InsturctPix2pix~\cite{arxiv:2022:instructpix2pix} utilized Prompt-to-Prompt to build a supervised dataset for image editing with natural language instruction and then finetuned a Stable Diffusion model with the dataset. However, these approaches still suffer from one of the following drawbacks. First, the user cannot constrain the region of editing, and editing methods often modify more pixels than it desires. Second, they do not resolve the issue that details are hard to describe in plain language and require trial and error to find the ideal editing instruction. 
 In this paper, we explore layered image generation guided by text with an intent of addressing the aforementioned challenges. As this is the first trial to generate image layers through diffusion models, we formulate the problem as generating a two-layer image, which includes foreground $F$, a background $B$, a layer mask $m$\footnote{We borrowed the terminology of the layer mask from Photoshop, which is a nondestructive way to hide parts of an image or layer without erasing them}, and the composed image. The layer mask controls a layer's level of transparency, and the composed image is formed based on alpha blending. Examples are shown in Figure~\ref{fig:dataset_examples}. With a layered image, a user is able to scale or move the foreground and replace foreground or background in a straightforward manner in image editing applications. Because of the layer mask, users' edits can be explicitly controlled to apply to either the foreground or background.

\begin{figure}[t!]
    \centering
    \begin{subfigure}{0.95\linewidth}
    \includegraphics[width=\linewidth]{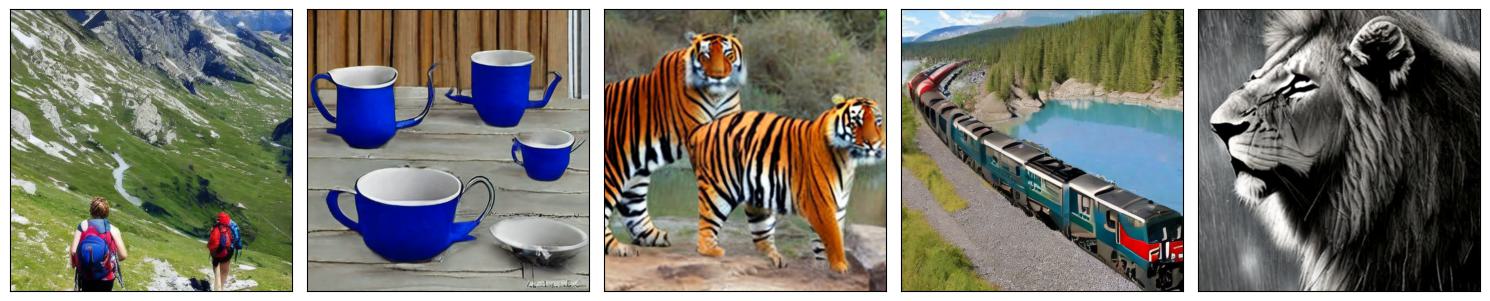}
    \end{subfigure}

    \begin{subfigure}{0.95\linewidth}
    \includegraphics[width=\linewidth]{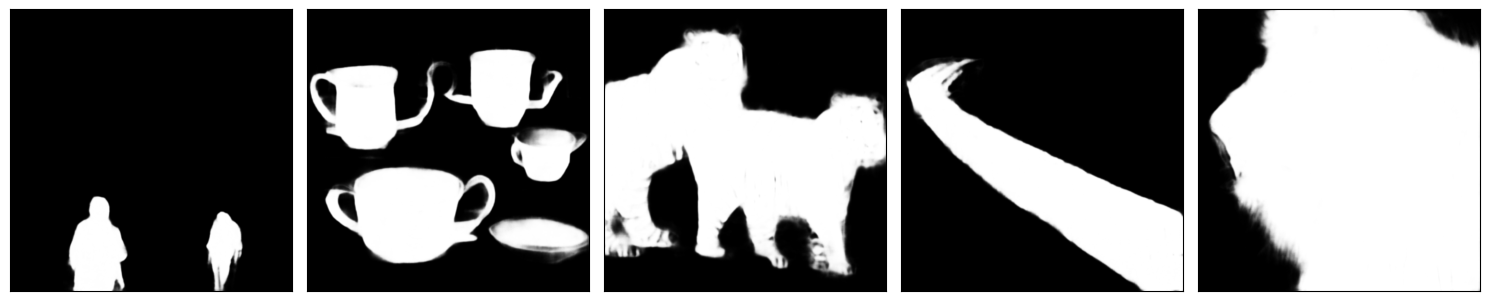}
    \end{subfigure}
    \caption{Layer mask examples. The scale, location, and number of objects vary largely.}%Composed images and masks from \aemodel-SD. \aemodel can generate two-layer images with diverse scales and a different number of foreground objects. We skip text prompts for this illustration. }
    \label{fig:scale_mulobj}
    \vspace{-0.15in}
\end{figure}
%\vspace{-0.1in}

The two biggest challenges are layered-image generation using diffusion models and training data synthesis. A trivial solution to generate layered images is running text2image multiple times, however, that doesn't guarantee the compatibility of the generated images in each run. Another solution is generating an image via a text2image system and then running salient object segmentation on the generated image. However, failure masks often occur due to the diverse scale and location of objects and the number of objects in the image. 

Motivated by the latent diffusion models, we train a layered-image autoencoder to learn their latent representation and then train diffusion models on the latent representation to generate layered images.
Concretely, the layered-image autoencoder learns to encode and decode two-layer images with a multi-task learning loss that includes reconstruction loss and loss terms inspired by image composition. We found that including the composed image as a layer in the autoencoder training further improves the generated image quality. %Second, we find that let CaT2I-AE auxiliary reconstruct the original images gives better generation performance. We guess the original images guide the model escaping from the noises introduced in the estimation of components.
Such layered output facilitates layer compositing and even improves the image synthesis quality. Meanwhile, the generated layer mask quality also benefits from the generative network in terms of adapting to different object scales as well as having multiple objects as a foreground as shown in Figure~\ref{fig:scale_mulobj}.

\begin{figure}
    \centering
    \begin{tabular}{lcr}
    \includegraphics[width=0.32\linewidth]{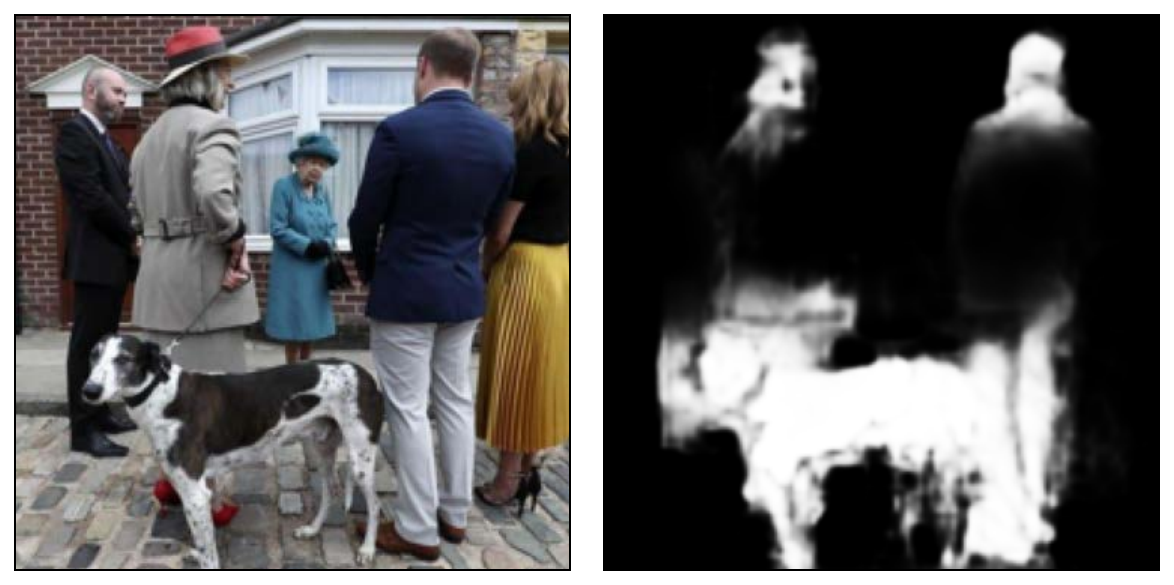}
    \includegraphics[width=0.32\linewidth]{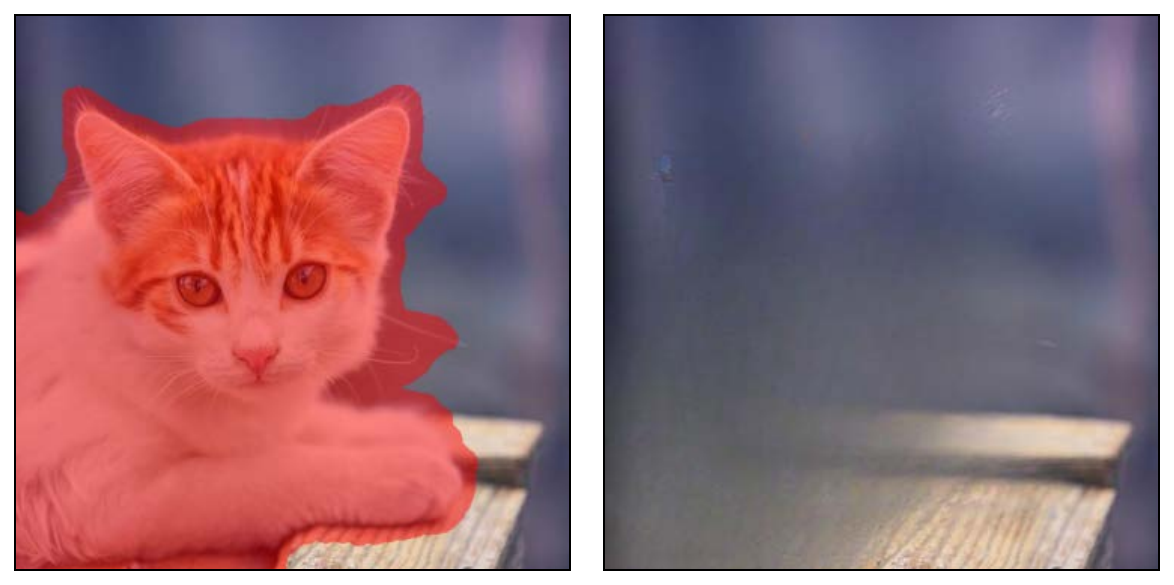}
    \includegraphics[width=0.32\linewidth]{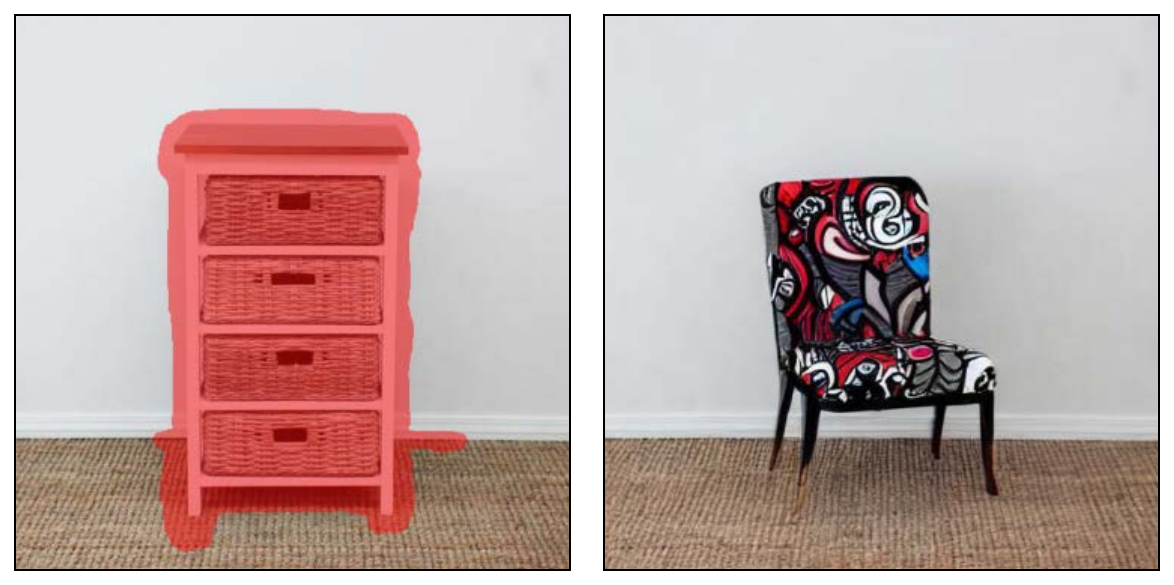}\\
    %Mask failure cases & \multicolumn{2}{c}{Inpainting failure cases}  
    \end{tabular}
    \caption{Failure cases of salient object segmentation (left) and inpainting (middle and right). The red shaded regions indicate the object to be removed. %\todo{remove the words in the image: image, mask, masked image etc... and adjust spacing of the 6 images.}
    }
    %ICON~\cite{tpami:2022:icon} predicts salient region of its input. LaMa~\cite{wcav:2022:lama} and Stable Diffusion V2 \cite{cvpr:2022:ldm} are image inpainting methods that fill the masked part~(red shaded region) of their input. } %Examples of failed to remove foreground objects in two state-of-the-art image inpainting methods: LaMa~(left, \cite{wcav:2022:lama}) and Stable Diffusion V2~(right, \cite{cvpr:2022:ldm}). 
    
    \label{fig:inpaint_failures}
    \vspace{-0.25in}
\end{figure}

To synthesize training data for layered image generation, we run salient object segmentation on training images used for text2image development. We chose to run salient object segmentation based on the observation that text prompts usually indicate the salient object clearly. As both the salient object segmentation and inpainting often fail (as shown in Figure~\ref{fig:inpaint_failures}), we developed an automatic data synthesis system that can distinguish high-quality salient object mask and high-quality impainting results. That system produces the proposed \dataset dataset, consisting of about $57.02M$ high-quality layered images among $600M$ LAION-Aesthetic dataset\footnote{\url{https://laion.ai/blog/laion-aesthetics/}}~\cite{nips:2022:laion5b}.
%We then run image inpainting to fill in missing content in the background. However, because the training data is extremely noisy, not every image has a clear salient object. Also, the state-of-the-art inpainting algorithms are not always working in complex situations\todo{need a figure. response: added as Figure~\ref{fig:inpaint_failures}}. To address these issues, 

%\dataset paves the way for text2layer generation. For the first time, we build a modified Stable Diffusion architecture for this task. In particular, we figured out directly ``re-use'' Stable Diffusion for text2layer led to undesirable results~(See Section~\ref{sec:eval}). To address the failure, we propose a simple yet effective layered-image autoencoder \aemodel to compress layers into a single latent vector, and then train a standard denoising diffusion model on the latent space. We incorporated two innovations in the \aemodel. First, \aemodel learns to encode and decode two-layer images with a multi-task learning loss that includes normal reconstructionl loss and loss terms inspired by image composition. Second, we find that let \aemodel auxiliary reconstruct the original images gives better generation performance. We guess the original images guide the model escaping from the noises introduced in the estimation of components.  

%\todo{innovations need better story telling ... }
As a first trial to generate layered images, we developed an evaluation metric to measure the output quality. Results showed that the proposed method generally produces layered images with better image quality in terms of FID, higher mask accuracy, and greater text relevance compared to several Stable Diffusion-inspired baseline models. %On image quality, our 512 \aemodel-SD even achieves a lower FID compared to SD-v1-4 (7.54 v.s. 9.13) on \dataset testset. For mask quality, \aemodel-SDs attain IOUs of 0.904 and 0.851 with the reference masks produced by a state-of-art salient detection model ICON and human annotations on models' predictions, respectively. The absolute gap with other baselines are about 0.125 and 0.086. About text-image relevance, {\aemodel}s are just slightly below SD-v1-4 in CLIP score, 0.251/0.261 for 256/512 resolution versus 0.279 for SD-v1-4, and outperforms all other baselines.
%our \aemodel-SD  

%Figure~\ref{fig:intro} displays several examples of failed editing. \todo{find some case} Therefore, users may perform final editing on the generated image with software or tools. On account of this observation, we present a new diffusion-based text-to-image generative model that is friendly to common image editing software like Adobe Photoshop\footnote{\url{https://www.adobe.com/products/photoshop.html}} and GNU GIMP\footnote{\url{https://www.gimp.org/}}. Specifically, our model generates a two-layer image with an alpha mask given user's prompt. 

%This work present the first efforts to .... In the future, .... 

%There are several challenges. 

%Results and comparisons ......

In summary, our contributions are three folds. 
% \vspace{-0.1in}
\begin{enumerate}
    \item We developed a text2layer method to generate a layered image guided by text, including a foreground, a background, a mask and the composed image.
    %\vspace{-0.05in}
    \item We introduced a mechanism to synthesize high-quality layered images for diffusion model training and produces $57.02M$ high-quality layered images for future research.
    %\vspace{-0.05in}
    \item We set a benchmark for layered-image generation. Experimental results showed that the proposed method generated higher-quality composed images with a higher text-image relevance score together with a better mask accuracy compared to baselines. 
    \vspace{-0.05in}
\end{enumerate}

\section{Related Work}
As layered image generation has not been systematically investigated, we discuss studies in text2image, text-based editing and image segmentation that are related to our work.

\vspace{2pt}
\noindent \textbf{Text-based image generation} Text-based image generation is the task of synthesizing images according to natural language descriptions. As a conditional generative problem, many previous works~(\cite{cvpr:attngan:2018,cvpr:xmcgan:2021}) approach this problem by training GANs~\cite{nips:gan:2014}) on image caption datasets. Another line of line work~(\cite{icml:2021:dalle,nips:2021:cogview}) tackles the generation in an auto-regressive way with Transformers~\cite{nips:2017:transformers}, usually utilizing VQ-VAE~\cite{nips:2017:vqvae} codebooks to produce image tokens. 
Recently, diffusion-based approaches significantly boost the generated image quality in terms of semantic coherence and fine spatial details, such as Imagen~\cite{arxiv:2022:imagen}, DALL·E 2~\cite{arxiv:2022:dalle2}, GLIDE~\cite{icml:2022:glide}, and Parti~\cite{arxiv:2022:parti}. Behind the scene, they are developed upon recent advances in denoising diffusion probabilistic models~(DDPM, \cite{nips:2020:ddpm, icml:2021:improved-dm}). DDPM~\cite{nips:2020:ddpm} and its variants~(\cite{iclr:2021:ddim}) learn the reversed process of a Markovian process that iterative adds noise to training images to produce white noise. Latent diffusion~\cite{cvpr:2022:ldm} trains the diffusion model on the latent space of a pretrained autoencoder instead of pixel space. For text2image generation, it incorporate classifier-free guidance~\cite{arixv:2022:clsfreeguide} to further improving sample quality and text-image relevance. We are motivated by the latent diffusion model and learn layered-image latent representations for layered image generation.

\vspace{2pt}
\noindent \textbf{Text-guided image editing} 
%Besides image generation, automatic image editing with textual guidance could help amateurs process their images conveniently and simplify some tedious work for professionals. Directly sampling from the above-mentioned text-based image generative models cannot fulfill the editing requirements since those models do not account for the objects and layouts in the original image and produce images utterly different from the original image. 
A line of papers~(\cite{iccv:styleclip:2021,arxiv:2021:stylegannada}) takes a pretrained CLIP~\cite{icml:2021:clip} image-text encoder to guide image manipulation. They usually measure the editing effect with cosine similarity between CLIP embedding of edited images and target texts. Style-CLIP~\cite{iccv:styleclip:2021} first inverts the input image into StyleGAN2's latent space and optimizes the latent vector with CLIP to find the target image. However, Style-CLIP and other GAN-based image editing methods~(e.g.,\cite{arxiv:2021:stylegannada}) need to solve challenging GAN inversion~\cite{eccv:2016:gan-inversion-1,iccv:2019:image2stylegan} problem first and do not support localized editing. Other works~\cite{eccv:2022:text2live,cvpr:2022:clipstyler} use CLIP without a pretrained generator by performing test-time optimization.

Inspired by the recent blooming of diffusion models, several works~(\cite{cvpr:2022:blended-diffusion,iclr:2022:sdedit,arxiv:2022:dreambooth,cvpr:2022:diffusionclip,arxiv:2022:imagic,arxiv:2022:prompt-to-prompt,arxiv:2022:instructpix2pix,iclr:2023:textual-inversion,arxiv:2022:sine}) propose to utilize a diffusion model as a generator for image editing. SDEdit~\cite{iclr:2022:sdedit} first adds noise to the input and then denoises the resulting images conditioned on a target. 
% DiffusionCLIP~\cite{cvpr:2022:diffusionclip} finetunes a pretrained diffusion model with CLIP similarity loss and deterministic DDIM sampling. 
Blended diffusion~\cite{cvpr:2022:blended-diffusion} performs localized editing with a region mask and a CLIP-guided inference algorithm for DDPM. To generate personalized images with text prompts, Textual Inversion~\cite{iclr:2023:textual-inversion} and DreamBooth~\cite{arxiv:2022:dreambooth} finetunes the text embedding of pretrained diffusion models to generate images of personalized objects in novel scenarios. Prompt-to-Prompt~\cite{arxiv:2022:prompt-to-prompt} modifies the cross-attention maps in the inference steps to manipulate images. SINE~\cite{arxiv:2022:sine} enables personalized image editing with a single image with model-based guidance and patch-based finetuning. InsturctPix2pix~\cite{arxiv:2022:instructpix2pix} creates a supervised model for image editing. %The authors collect a human-annotated text-driven image editing instruction dataset, augment the dataset with GPT3, and then get the ground-truth with \cite{arxiv:2022:prompt-to-prompt}. 
Imagic~\cite{arxiv:2022:imagic} combines latent space interpolation and model finetuning to edit images with a text prompt. Direct Inversion~\cite{arxiv:2022:direct-inversion} claims that simply denoising the latent vector of the input image with text instruction produces satisfying editing. 
Most of these approaches cannot constrain the region of editing, and we aim to generate explicit salient masks together with the image to facilitate editing and layer compositing workflows.

\vspace{2pt}
\noindent \textbf{Image matting} Image matting studies extracting the foreground alpha matte $\alpha$ from an image $I$, which is useful for image composition and editing. Formally, image matting algorithms solve an alpha matte $\alpha$, such that for the $i$-th pixel of $I$: 
$I_i \approx \alpha_i F_i + (1 - \alpha_i) B_i$, where $F$ and $B$ are unknown foreground and background. We take the same formulation to compose the foreground and background in the layered image generation.

Image matting have been extensively studied~\cite{tpami:2007:closedform,cvpr:2017:deepmat,cvpr:2021:sim,ijcv:2022:gfm}. Traditional approaches~\cite{tpami:2007:closedform,cvpr:2011:globalmat} for matting rely on affinity or optimization. DIM introduced Composition-1k dataset, and took deep convolutional networks for image matting. Since then, many new deep learning methods for image matting have been proposed~\cite{cvpr:2021:sim,aaai:2020:gca-matting,aaai:2022:modnet}. Recently, MatteFormer~\cite{cvpr:2022:matteformer} attacked matting with a modified SWIM Transformers~\cite{cvpr:2021:swin}. Motivated by the success of matting methods, we leverage composition loss~\cite{cvpr:2017:deepmat} and Laplacian loss~\cite{cvpr:2019:ctx-mat} to train our autoencoder.  

Salient object segmentation/detection (SOD) aims at highlighting visually salient object regions in images~\cite{tpami:2021:sod-dl}. Similar to image matting, deep architectures~\cite{tpami:2022:icon,cvpr:2015:sod-mdf,cvpr:2017:sod-dss,aaai:2020:sod-global} beat the performance of traditional methods~\cite{tpami:2014:trad-sod-1,ejn:2003:trad-sod-2}.
DSS~\cite{cvpr:2017:sod-dss} proposed CNN with short-cut connections between the deeper and shallower features for SOD. To deal incomplete saliency, 
%TSPOA~\cite{iccv:2019:sod-tspoa} mines part-object relationships and employs such relationships with Capsule Network~\cite{nips:2017:dynamic-routing}. 
GCPANet~\cite{aaai:2020:sod-global} built a progressive context-aware feature aggregation module for effective feature fusion in SOD. 
ICON~\cite{tpami:2022:icon} is a deep architecture that improves the integrity of the predicted salient regions. Higher integrity of composition masks is vital to build our dataset since composition with an incomplete mask will tear apart in the foreground and background and produce many artifacts. Therefore, we chose ICON for estimating masks.

\section{Synthesizing High-Quality Layered Images}

\label{sec:dataset}

We formally define the two-layer image and introduce the proposed approach for high-quality layered image synthesis in this section. With the proposed approach, we generated a $57.02M$  high-quality layered-image dataset, named \dataset, which includes $57M$ for training and $20K$ for testing. We refer to the test set as LL\textsuperscript{2}I-E in the rest of this paper. %\revise{For the purpose of empirical studies, we randomly sample 20000 subset from the testing set, and refer to it as LL\textsuperscript{2}I-E for the rest.  }
%\todo{[SINCE WE ARE USING 20K FOR TESTING, NEVER USED 0.6M. DO WE STILL NEED TO MENTION 0.6M HERE?]}
%We start this section with a formal definition of a two-layer image and treat its generation task as the central theme of this paper. We develop a two-step solution to the problem. The first step is building a large-scale dataset consist of two-layer images with text descriptions. We name the dataset \dataset. We introduce details on its construction and discuss its relationship with existing datasets in the literature.

\subsection{Definition of Two-Layer Image} 
\label{subsec:task}
%\vspace{2pt}
%\textbf{Two-layer Image}. 
Intuitively, a two-layer image represents an image with a foreground and a background, as well as a mask $m$ to composite the two layers. Formally, a two-layer image is a triplet $\mathcal{I} = (F, B, m)$ where $F$, $B$, and $m$ are the introduced foreground, background and mask, respectively. Throughout the paper, we assume they are of the same spatial dimension $H \times W$. Hence, $F, B \in \mathbb{R}^{3 \times H \times W}$ and $m \in \mathbb{R}^{H \times W}$. Figure~\ref{fig:dataset_examples} displays samples from \dataset dataset. %As one will see, we've made multiple efforts to supply two-layer images with high-quality foreground/background components and accurate masks.

\subsection{Overview of \dataset Dataset Construction}
As shown in recent text2image efforts, large-scale datasets such as LAION-2B and LAION-5B~\cite{nips:2022:laion5b} are essential for training diffusion models~\cite{cvpr:2022:ldm}. To generate pairs of two-layer images $\mathcal{I}$ and text prompts, we propose an automatic data annotation method to create a two-layer image out of each image in the LAION dataset, so that they are of a large scale and paired with text prompts. In particular, to cope with limited computational resources, as a first trial, we chose to generate layered images using the LAION-Aesthetics, (short for LAION-A) subset, which contains around 600 million images with high aesthetic quality and text relevance\footnote{\url{https://laion.ai/blog/laion-aesthetics/}}. 

To generate a two-layer image out of each image in LAION-A, 
%We construct our \dataset dataset on top of the public LAION-5B~\cite{nips:2022:laion5b} dataset. The LAION-5B dataset consists of more than five billion CLIP-filtered image-text pairs. For our purpose, we take the LAION-Aesthetics~(short for LAION-A) subset, which includes \todo{600} million images with higher aesthetic quality and text relevance. Our dataset creation is quite intuitive. 
we apply a salient object segmentation method to extract the foreground parts from the original images and then fill the missing regions of the backgrounds using state-of-art image inpainting techniques. We chose salient object segmentation instead of segmenting objects of specific semantic categories for two purposes. First, it sets the promise of being able to generalize to a large-scale dataset. Second, the text prompt associated with each image is generally applicable to a two-layer image that is composed of a salient object, background, and a mask. Meanwhile, as salient object detection and inpainting methods do not always produce high-quality output, we trained two classifiers to filter out the samples that are with bad salient masks or bad inpainting results. 

\subsection{Extracting foreground and background parts}

We curated a two-layer image dataset based on LAION-A on the resolution of $512\times512$. As we defined a two-layer image $\mathcal{I}$ as a triplet of $(F, B, m)$, we take a straightforward way of estimating the triplet from an existing image $I$ such that
\begin{equation}
    I \approx m F + (1 - m) B
    \label{eq:comp}
\end{equation}
Eq~\eqref{eq:comp} is under-constrained with 6 degrees of freedom for each pixel. We estimate them as follows. 

\vspace{2pt}
\noindent \textbf{Estimation of $F$ and $m$}.
 In this work, we predict salient mask $m$ from $I$ using state-of-the-art salient object detection method ICON~\cite{tpami:2022:icon}. For the foreground image $F$, we set a threshold~(0.1 for our case) for $m$ and directly copy pixel values $I_i$ of $I$ at spatial locations which have mask value $m_i$ greater than the threshold. 
 %\todo{In principle, any image matting, instance segmentation or salient object detection techniques could be used for extracting a region mask $m$ from $I$ denoting the foreground region.}

\vspace{2pt}
\noindent \textbf{Estimation of $B$}.
We utilize image inpainting technique to acquire the background $B$ for an image $I$. Concretely, we first apply a dilation operation on $m$ to obtain an augmented $\tilde{m}$, which helps alleviate the errors in the estimation of $m$. Next we use the state-of-art diffusion-based inpainting\footnote{\url{https://huggingface.co/stabilityai/stable-diffusion-2-inpainting}} to fill the masked region and produce the inpainted image $B$. We attempted to use the prompts ``background'' to enhance the background generation. However, we did not observe quality improvement in the output. So we chose to feed empty text prompt into the inpainting model.

% \todo{some comments ... (1) define unfiltered dataset in Table 4 here; (2) describe the dataset statistics, such as a histogram of salient object size.}

In Section~\ref{sec:quality_filter}, we will introduce how we further filter the data to obtain a higher quality dataset in detail. 
To differentiate from the filtered dataset, we name the originally built dataset \dataset~(U), where U means unfiltered. Please refer to supplementary material for more statistics of the \dataset. Some examples from \dataset are visualized in Figure~\ref{fig:dataset_examples} (more in the supplement).
% Similar to $\alpha$ and $F$, we utilize a data-drive approach to filter out low-quality or undesired background inpainting.  

\subsection{Quality filtering}
%\todo{introduce the criteria of manual labeling}
%\todo{summarize dataset statistics post filtering, how many for training and how many for testing.} 
%{\color{blue} The whole sub-section is re-organized as follow.}
\label{sec:quality_filter}

\begin{figure}[t]
    \centering
    \begin{subfigure}{0.45\textwidth}
        \includegraphics[width=\linewidth]{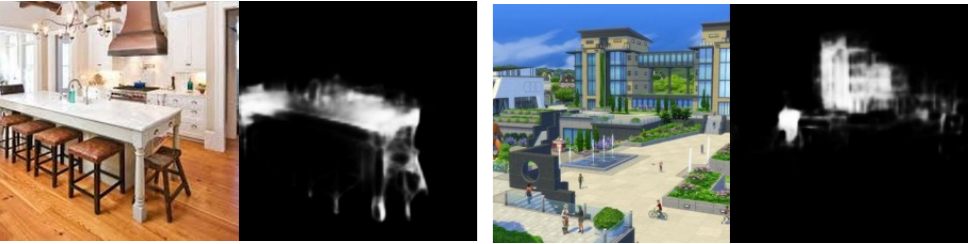}
        \caption{Predicted ``bad'' masks}
    \label{fig:salient_quality_before}
    \end{subfigure}
    \begin{subfigure}{0.45\textwidth}
        \includegraphics[width=\linewidth]{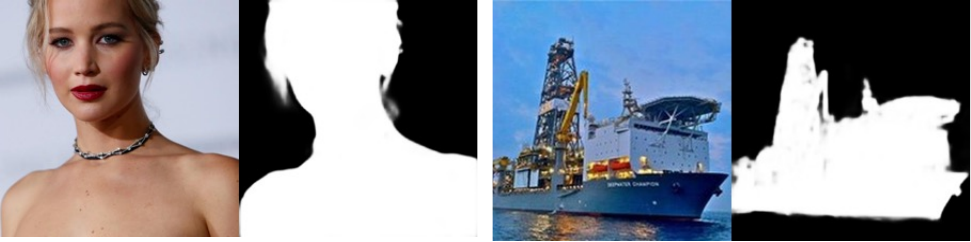}
        \caption{Predicted ``good'' masks}
    \label{fig:salient_quality_after}
    \end{subfigure}
    \begin{subfigure}{0.45\textwidth}
        \includegraphics[width=\linewidth]{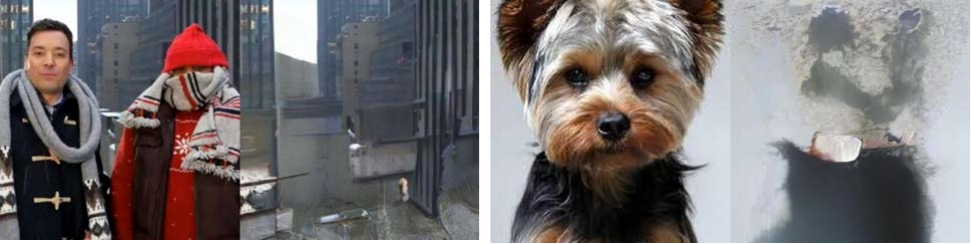}
        \caption{Predicted ``bad'' inpaintings}
    \label{fig:inpainting_quality_before}
    \end{subfigure}
    \begin{subfigure}{0.45\textwidth}
        \includegraphics[width=\linewidth]{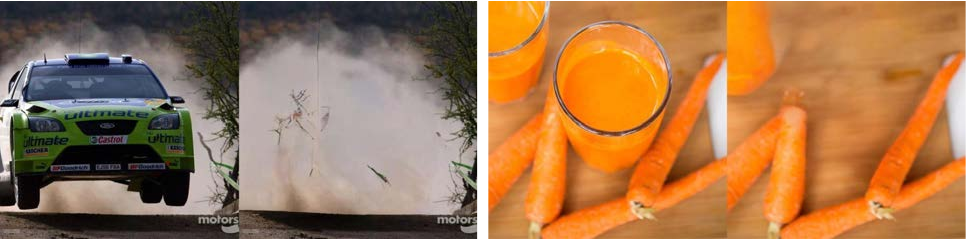}
        \caption{Predicted ``good'' inpaintings}
    \label{fig:inpainting_quality_after}
    \end{subfigure}
    
    \caption{Predicted good and bad salient masks and inpaintings}
    \label{fig:salient_quality}
\vspace{-0.2in}
\end{figure}
We noticed that the mask prediction model for $\alpha$ and inpainting model for $B$ do not always produce desired results. Specifically, as shown in Figure~\ref{fig:salient_quality_before}, some of the masks are incomplete or contain too many background elements. As shown in Figure~\ref{fig:inpainting_quality_before}, the inpainting model may introduce visual artifacts. In addition, the inpainting model tends to retrieve the objects in the missing regions. However, we expect  it to only patch the background instead of the foreground. Otherwise, the diffusion model tend to generate contents that are not relevant to the text descriptions.

To solve the aforementioned issues, we train two classifiers to remove the samples that are with bad saliency map or inpainting quality. We first manually annotated 5000 training samples and 1000 test samples respectively. Specifically, we randomly sampled the generated two-layer images from \dataset(U). For salient mask quality labeling, a sample is labeled as a bad sample if one of the three conditions occur: 
% \vspace{-0.1in}
\begin{itemize}
    \item The object masked out is largely incomplete.
    % \vspace{-0.1in}
    \item The mask contains regions from the backgrounds.
    % \vspace{-0.1in}
    \item The selection of the salient object is too small
\end{itemize}
For inpainting quality labeling, a sample is labeled as a bad sample if there is an object or artifacts in the inpainted region or the inpainted region is not cohesive with the remaining image. Typical failure cases are shown in the Figure~\ref{fig:inpaint_failures}.

To obtain the two quality filters as mentioned above, we train two different classifiers with the same  neural network architecture and training schedules. For mask quality classifier, the inputs are the original images $I$ and their corresponding masks $m$. The inputs to the inpainting quality classifier are the inpainted images. The output of both classifiers are the quality labels. For implementation, we replace the last layer of EfficientNet~b0~\cite{icml:2019:efficientnet} with a fully connected layer with one single neuron which indicates the probability of a sample is good or not. Then we only fine-tune the last layer while freeze all other layers. After the classifiers are well-trained after 100 epochs, we feed all the samples from \dataset(U) into the two classifiers and discard the samples with bad masks or poor inpainting results, which results in a filtered dataset \dataset which contains $57.02M$ samples. The filtered results are visualized in Figure~\ref{fig:salient_quality_after} and Figure~\ref{fig:inpainting_quality_after}.
More statistics and analysis of \dataset and data filtering are available in the supplement.

\section{Modeling}

% \begin{figure}[t!]
%     \centering
%     % \includegraphics{}
%     \caption{A figure for model}
%     \label{fig:modelling}
% \end{figure}

%\vspace{2pt}
\subsection{Text2Layer Formulation}
Our main task is to design a model which can generate a two-layer image $\mathcal{I}$ given a text prompt. 
Formally, we train a conditional generative model to model the probability distribution $p_\theta(\cdot;\cdot)$ such that 

\begin{equation}
    \mathcal{I} = (F, B, m) \sim p_\theta(z; T)
\end{equation}
where $T$ is the text prompt and $\theta$ is the parameter of the distribution $p$. 

% Our solution for this task is building a dataset $\mathcal{D}$ of $(\mathcal{I}, T)$ pairs and then train a novel latent-space diffusion model on this dataset. The dataset \dataset includes 50 million (\todo{Is 50 accurate?}) two-layer images with \todo{satisfying} quality. On the model side, to compress two-layer images into a compact vectors for a latent-space diffusion model, we extend an autoencoder for RGB image to two-layer image space with \todo{several} \todo{innovations v.s. modifications}. We name the autoencoder \aemodel and our full model as \aemodel-SD.

Equipped with \dataset dataset, we are ready to build a conditional diffusion model to synthesize a two-layer image $\mathcal{I}$ given text $T$. Our model architecture is based on Stable Diffusion~\cite{cvpr:2022:ldm}, which has already demonstrated high-quality results in text2image generation. 

We need to adapt Stable Diffusion to the two-layer image generation task. %\sout{Recall that Stable Diffusion applies DDPM~\cite{nips:2020:ddpm} training on the latent space of a pretrained image autoencoder with a variant of UNet~\cite{miccai:2015:unet}.}
Recall that Stable Diffusion employs a pre-trained autoencoder to project the original images to a latent space in a significantly lower dimension before feeding them to a noise prediction module which takes the form of a UNet~\cite{miccai:2015:unet} variant.
%\sout{We can either modify the autoencoder part or the UNet part. We take the former approach in this paper, and we present another of our key innovations in the rest of this section -- a novel autoencoder architecture tailored for our two-layer image format, termed Composition-Aware Two-Layer Autoencoder~(\aemodel). We refer to the autoencoder of Stable Diffusion for conciseness as SD-AE.}
In our scenario, we found that in addition to reducing computational cost, the autoencoder also plays an essential role in capturing the inherent structure of two-layer image $I$. Therefore, we designed a novel architecture for autoencoder, termed Composition-Aware Two-Layer Autoencoder~(\aemodel), tailored for two-layer image generation which can be easily generalized to multiple layers. In Section~\ref{sec:results}, we demonstrate that such a simple design is sufficient to produce high-quality two-layer images.

\subsection{\aemodel Architecture}

%\todo{To decorate the naive architecture ...}
In the original SD-AE~(the autoencoder used in Stable Diffusion), the decoder reconstructs an input image $I$ with the encoder's output latent vector $z$. We detail the changes made in \aemodel, which compresses two-layer images into latent space and reconstructs them from the latent vectors. For the encoder, we stack $F$, $B$, and $m$ into a single tensor $X$. 
%\todo{leave concrete number to implementation details as in some of the exp the number is 7 and some the number is 10 response: updated}. 
We keep other parts of the encoder intact. While a complicated encoder is possible, we find this approach already achieved good generation results.

%\sout{For the decoder, we distinguish the generation of image components and mask component.} 
To obtain a better decoded result, we differentiate the generation process of image components and mask component for the last layer by applying multiple prediction heads. Specifically, we remove the last layer of SD-AE decoder and attach three prediction heads to predict $F$, $B$ and $m$, respectively. Each prediction head is a tiny two layer convolutional network of \textit{conv-bn-relu-conv}. Another modification we made compared to SD-AE is that we added an auxiliary output branch to predict the original image $I$. Our motivation is the supervision of the original image $I$ introduced extra information in addition to $F$ and $B$. In Section~\ref{sec:results}, we show that this extra supervision slightly improves the generation quality. %and it is less noisy than the estimated two-layer images $\mathcal{I}$.  

\subsection{Training Objective}
%\textbf{[TODO DIFFUSION TRAINING OBJECTIVE]} 
Our training objective for \aemodel extends the original autoencoder training objective in \cite{cvpr:2022:ldm} with terms for masks. Given the autoencoder's reconstructed $(\hat{F}, \hat{B}, \hat{I}, \hat{m})$, the full multi-task loss $L$ is defined as
\begin{equation}
    L = L_\mathrm{img}(\hat{F}, \hat{B}, \hat{I}; F, B, I) + \lambda L_\mathrm{mask}(\hat{m}, m)
\end{equation}
where $L_\mathrm{img}$ denotes the loss for image components~($B, F, I$), $L_\mathrm{mask}$ denotes the loss for mask channel, and $\lambda$~(we take 1 in all the experiments) controls the tradeoff of reconstruction quality between image components and mask component. As for the image components $L_\mathrm{img}$, we directly use a combination of LPIPS~\cite{eccv:2018:lpips}, $\ell_1$ norm and adversarial loss as in SD-AE. For the mask $m$, motivated by the works~\cite{eccv:2021:mgm} on image matting, we consider a combination of $\ell_1$ loss, Laplacian loss~\cite{cvpr:2017:deepmat}, and composition loss~\cite{cvpr:2019:ctx-mat} (details in the supplement). 
\begin{equation}
    L_\mathrm{mask} = \ell_1(m, \hat{m}) + 2 \ell_\mathrm{comp}(m) + 3 \ell_\mathrm{lap}(m, \hat{m})
\end{equation}
%\todo{See supplement material for the details}

After \aemodel is well trained, we utilize the same cross-modal attentive UNet architecture used in work \cite{cvpr:2022:ldm} to train a DDPM on the latent space of \aemodel. In particular, let $f_\theta$ be the UNet, and the loss function $L_\mathrm{DDPM}$ is defined as
\begin{equation} 
    L_\mathrm{DDPM} = E_{z, \epsilon, t} \left[ \left\| \epsilon - f_\theta(z_t, t) \right\|_2^2  \right]
\end{equation}
where $t$ is a random time step between 1 and $t_\mathrm{max}$, and $\epsilon$ is an independent Gaussian noise, $z$ is the latent vector produced by \aemodel, and $z_t$ is a noisy version of $z$. 
% We simply call our full model as \aemodel-SD.

% \subsection{Training Details}

\section{Experiments}
\label{sec:eval}

In this section, we extensively compare the proposed text2layer method on the two-layer image generation task to several baselines on the \dataset dataset introduced in Section~\ref{sec:dataset}. The evaluation focuses on the image quality, mask quality and text-image relevance. % We then close this section with ablation studies on our proposed method. 

% \subsection{Datasets and Baselines}

% \vspace{2pt}
% \textbf{Datasets}.

% MSCOCO dataset~\cite{eccv:2014:mscoco}

% \vspace{2pt}
% \textbf{Baselines}. 

\subsection{Baseline Methods}

We name our full model \aemodel-SD, and compare it with the following baseline methods on our \dataset dataset. 
\begin{itemize}
    \item SD-v1-4 and SD-\dataset: SD refers to the Stable Diffusion model for text-to-image synthesis. We compare the quality of \aemodel-SD's composed images $I = m F + (1 - m) B$ with images generated from Stable Diffusion v1-4\footnote{\url{https://huggingface.co/CompVis/stable-diffusion-v1-4}}. Additionally, because the \dataset dataset is significantly smaller than the LAION-A dataset, we trained a SD model\footnote{We re-used the SD-AE and trained the same UNet architecture from scratch for the diffusion model as SD does.} for text2image using the \dataset dataset for fair comparison. We denote these two SD models as SD-v1-4 and SD-\dataset.
    % \xl{which version of SD? include the link in the footnote} 
    % \item SD-\dataset: we follow the training approach of Stable Diffusion 
    \item SD-AE-UNet and SD-AE-UNet~(ft): for these two baselines, we pass each component of $\mathcal{I}$~(a.e.,$F, B, m$) into the SD-AE, and train a UNet-archtecture diffusion model which takes the stacked latent vectors as inputs. Formally, let $g(\cdot)$ be the SD-AE, we feed $z$ defined as below into the UNet. 
    \begin{align}
        & z_F, z_B, z_m= g(F), g(B), g(m) \nonumber \\
        & z = \mathrm{concat}_\mathrm{ch}(z_F, z_B, z_m)
    \end{align}
    where $\mathrm{concat}_{ch}(\cdots)$ concatenates input tensors along the feature channel.
    For SD-AE-UNet, we train the UNet model from scratch, while for SD-AE-UNet~(ft), we finetune the UNet from the SD-v1-4. By evaluating \aemodel-SD against these two baselines, we show that it is essential to have a well-designed autoencoder for two-layer image generation task. 
    \item \aemodel-SD~(no-sup): this is a variant of \aemodel-SD without the supervision branch for the original image $I$. We show the effectiveness of introducing this additional supervision branch in Section~\ref{sec:results}. 
    
\end{itemize}
Because training high-resolution diffusion models (i.e., $512\times512$) requires excessive computational resources, we primarily evaluate all methods on the $256 \times 256$ resolution. In particular, all the methods except SD-v1-4 target at a resolution of $256$. We also trained a \aemodel-SD on the resolution of $512$, named \aemodel-SD~(512) to demonstrate the generation quality of various resolutions. %in generating higher-resolution two-layer images, we train a 512 version of \aemodel-SD, named \aemodel-SD~(512). %Since we do not have access to a well trained SD-v1-4 checkpoint which produces $256\times256$ images directly, we downsample the $512\times512$ images generated by SD-v1-4 to $256\times256$.\todo{which notation refers to 256-SD-v1-4?} 
% \todo{summarize the resolution of these approaches used in comparisons}

\subsection{Metrics}
\label{subsec:metrics}
To quantitatively analyze the capability of \aemodel-SD in synthesizing two-layer images in an extensive way, we take the following metrics. 
%\todo{simplify wording: FID, CLIP -- make it clear that we follow the same metric used in other text2image literature, focus the the dataset differences. IOU: need to complete. response: addressed}

\vspace{2pt}
\noindent \textbf{Fréchet inception distance~(FID)} FID is a metric used to evaluate the quality of images generated by a generative model~\cite{nips:2017:fid}. It compares the distribution of generated images with the distribution of real images. 

We follow the same FID metric as done in \cite{cvpr:2022:ldm,arxiv:2022:imagen} and take FID to measure the fidelity of composed images. We leverage the Clean-FID proposed in \cite{cvpr:2022:clean-fid}\footnote{\url{https://github.com/GaParmar/clean-fid}}. As the FID score significantly depends on the real image distribution, we report FID scores on two test sets. 
% \sout{The major one is the $20K$ random testing subset, LL\textsuperscript{2}I-E~(See Section~\ref{sec:dataset}). Plus, we also quantify models ability to generate diverse images beyond \dataset with a $20K$ unfiltered subset of LAION-A, which does not overlap with \dataset. We abbreviate this one \textit{LA} in experiments.}
One is LL\textsuperscript{2}I-E~(See Section~\ref{sec:dataset}) which contains $20K$ test samples. The other is a subset from unfiltered LAION-A, which also has $20K$ samples. The two sets are mutually exclusive. We abbreviate the latter one \textit{LA} in the following experiments.  We introduce \textit{LA} for quantifying our model’s ability to generate diverse images beyond the distribution of \dataset.

\vspace{2pt}
\noindent \textbf{CLIP score}
Previous work~\cite{emnlp:2021:clipscore} discovered that the CLIP~\cite{icml:2021:clip}, a vision-language model pre-trained on $400M$ noisy image and caption pairs, could be used for automatic evaluation of image captioning without the need for references. In particular, the cosine similarity between the features of the image encoder and language encoder could serve as a surrogate metric for image-text relevance. We follow the same CLIP score metric as it was used in \cite{iclr:2022:clipdraw,arxiv:2021:vqdiffusion} and refer to the cosine similarity as CLIP score. In particular, we take the CLIP score to validate that images composed by our \aemodel-SD faithfully reflect text prompts. We calculate the CLIP score on the LL\textsuperscript{2}I-E subset with Vit-L/14. 

\vspace{2pt}
\noindent \textbf{Intersection-Over-Union~(IOU)}.
IOU is widely adopted to assess the performance of % object detection, 
semantic segmentation.
% and instance segmentation methods. 
It computes the overlapping between a pair of prediction mask and ground-truth mask % 
% ~(or segmentation/instance masks) 
as a fraction.

In addition to the quality of composited images, we expect masks $m$ to capture meaningful regions in the composited images. To evaluate the generated masks, we compute the IOU of models' predicted mask $\hat{m}$ with two sets of ``reference'' masks because no ground-truth masks are available for the generated masks given certain text prompts. We use ICON's~\cite{tpami:2022:icon} predictions on the composed images as the first reference set, since that is how we estimate masks in dataset construction. For the second set, we manually annotate the masks of a test set which consists of 1500 generated composited images for each model. 

\subsection{Implementation Details}
All the experiments are running on four 8$\times$A100 GPU machines. For SD-AE-UNet and SD-\dataset, we train the UNet model for 75k steps with DDPM training. For SD-AE-UNet~(ft), we expand the number of input channels and output channels for the UNet model and initialize the weight from SD-v1-4, we then finetune the UNet model for 25k steps. \aemodel-SD and \aemodel-SD~(no-sup) require training the autoencoder from scratch, and we train both of them for 250k steps. We follow the same training scheme on UNet for both models as the SD-AE-UNet. To train the 512 resolution \aemodel-SD~(512), we resume from \aemodel-SD weight and train for 75k steps with 512 images.
%\todo{move resolution discussion to 5.1 and clarify that SD-v1-4 is 512 which we don't have a 256 checkpoint. Include the resolution info in the tables 1 and 3. response: addressed}
During inference, we sample two-layer images with DDIM~\cite{iclr:2021:ddim} scheme and classifier-free guidance~\cite{arixv:2022:clsfreeguide}. In particular, the number of DDIM iterations is $50$ and the guidance scale is $7.5$.   

\begin{figure}
    \centering
    % \begin{subfigure}{\linewidth}
    %     \includegraphics[width=0.\linewidth]{figures/generations/images/ours_10ch_fig_2.jpg}
    %     \hspace*{\fill}
    % \end{subfigure}
    \begin{subfigure}{0.47\linewidth}
        \includegraphics[width=\linewidth]{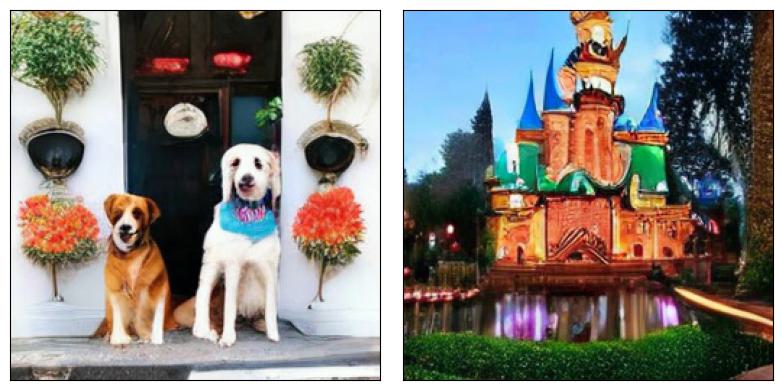}
    \end{subfigure}
    \begin{subfigure}{0.47\linewidth}
        \includegraphics[width=\linewidth]{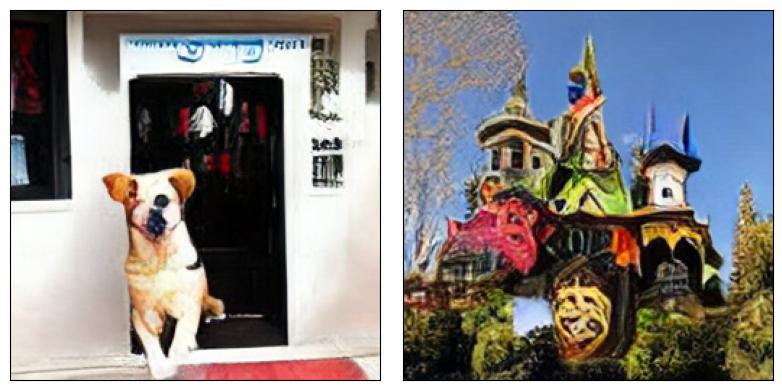}
    \end{subfigure}
    \begin{subfigure}{0.47\linewidth}
        \includegraphics[width=\linewidth]{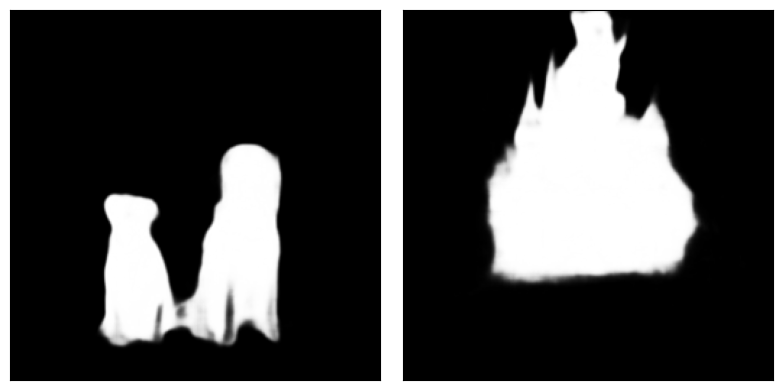}
        \caption{\aemodel-SD}
    \end{subfigure}
    \begin{subfigure}{0.47\linewidth}
        \includegraphics[width=\linewidth]{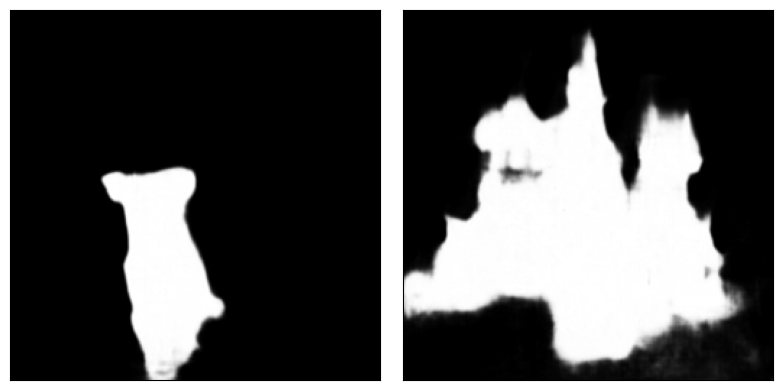}
        \caption{\aemodel-SD~(no-sup)}
    \end{subfigure}
    \begin{subfigure}{0.47\linewidth}
        \includegraphics[width=\linewidth]{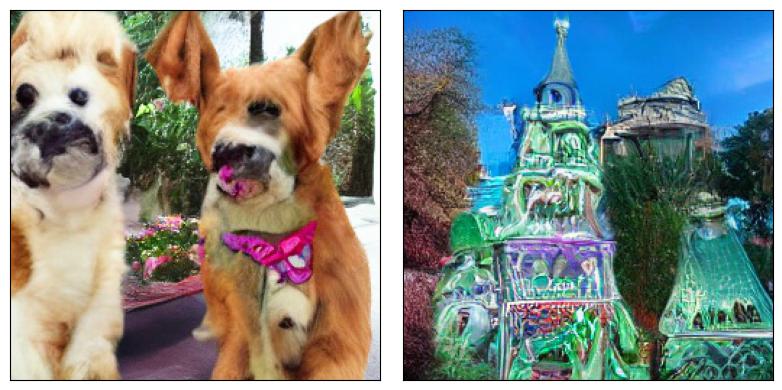}
    \end{subfigure}
    \begin{subfigure}{0.47\linewidth}
        \includegraphics[width=\linewidth]{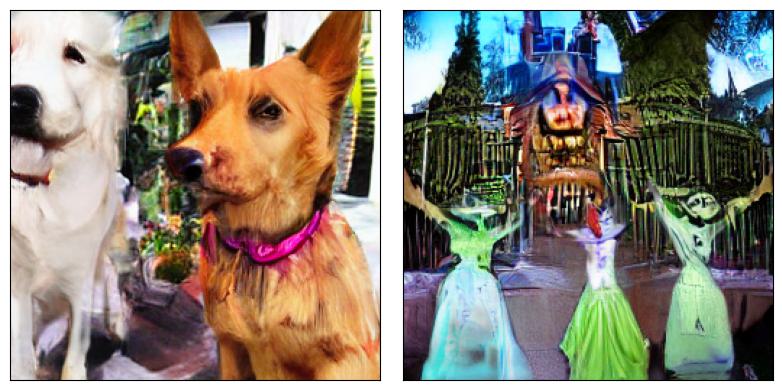}
    \end{subfigure}
    \begin{subfigure}{0.47\linewidth}
        \includegraphics[width=\linewidth]{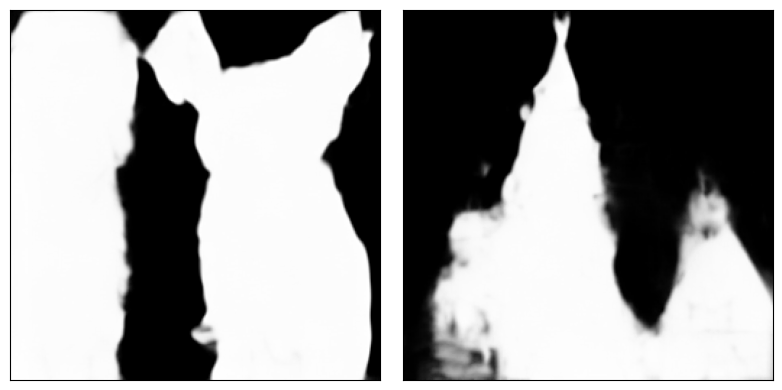}
        \caption{SD-AE-UNet}
    \end{subfigure}
    \begin{subfigure}{0.47\linewidth}
        \includegraphics[width=\linewidth]{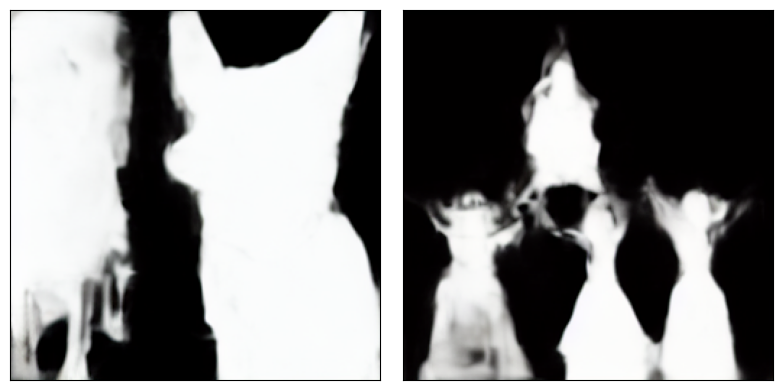}
        \caption{SD-AE-UNet~(ft)}
    \end{subfigure}
    \caption{Composited samples from \aemodel-SD and baseline models for $256 \times 256$ resolution. The prompts are (1) \textit{Dogs at the 
entrance of Arco Iris Boutique} and (2) \textit{Haunted Mansion Holiday at Disneyland Park}. Each $2 \times 2$ block displays the composited images and masks $m$ of the corresponding models. Find more in the supplement. }
    \label{fig:samples}
    \vspace{-0.15in}
\end{figure}
\subsection{Qualitative Results}
 Figure~\ref{fig:samples} displays samples from \aemodel-SD and baseline models for the $256\times256$ results. As shown in the figure, the samples from \aemodel-SD clearly have better details with sharper and more accurate masks compared to other methods. On composited images, \aemodel-SD's results have fewer artifacts and distortion. On masks, \aemodel-SD's results are more accurate, while results generated by SD-AE-UNet and SD-AE-UNet~(f) are noisy. 

\begin{figure}[t!]
    \centering
    \includegraphics[width=0.95\linewidth]{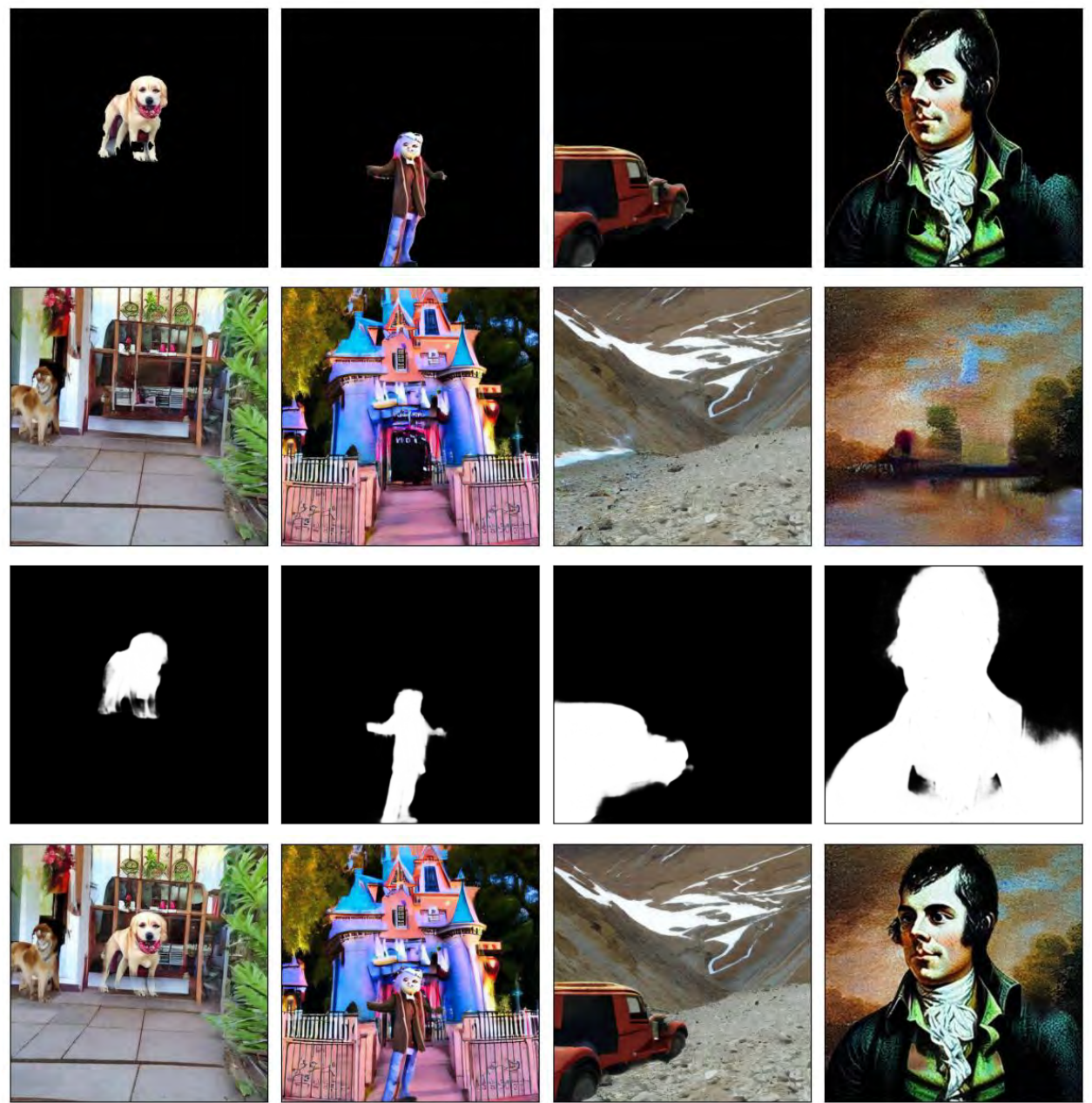}
    \caption{$F, B, m$ components and composed images for two-layer images sampled from \aemodel-SD~(512). Prompts are omitted. From top to down: $F$, $b$, and $m$. For more samples \textit{cf.} the supplement.} 
    \label{fig:samples_512}
    \vspace{-0.1in}
\end{figure}

% Please add the following required packages to your document preamble:
% \usepackage{multirow}
\begin{table*}[h!]
\centering
\begin{tabular}{c|c|c|c|c|c}

Dataset                                  & Resolution           & Method                                             & FID~(LL\textsuperscript{2}I-E, $\downarrow$) & FID~(LA, $\downarrow$)                   & CLIP Score~($\uparrow$) \\ \hline
\multirow{5}{*}{\dataset} & \multirow{5}{*}{256} & \aemodel-SD                         & \textbf{10.51}                      & \textbf{14.83} & \textbf{0.251}      \\
                                         &                      & \aemodel-SD (no-sup)                & 13.84                                                & 19.54                           & 0.241      \\
                                         &                      & SD-AE-UNet                                         & 18.53                                                & 21.96                           & 0.219      \\
                                         &                      & SD-AE-UNet (ft)                                    & 20.15                                                & 24.84                           & 0.234      \\
                                         &                      & SD-\dataset                         & 11.80                                                & 15.06                           & 0.250      \\ \hline
\dataset                  & \multirow{2}{*}{512} & \aemodel-SD  (512)                  & \textbf{7.54}                       & 13.01                           & 0.261      \\
LAION-2B + LAION-A                       &                      & SD-v1-4~\cite{cvpr:2022:ldm} & 9.13                                                 & \textbf{9.13}  & \textbf{0.279}     
\end{tabular}
\caption{ FID and CLIP scores for our \aemodel-SD and various baselines. FID is evaluated  on LL\textsuperscript{2}I-E and LA testset, and CLIP score is tested on LL\textsuperscript{2}I-E. The arrows indicate preferred directions for the metrics. 
\label{tab:fid_clip}}
\vspace{-0.1in}
\end{table*}

\begin{table}[h!]
    \centering
    \begin{tabular}{c|c|c}
        Model  & IOU~(ICON) & IOU~(human) \\
        \hline
       \aemodel-SD  &  0.885  &  0.799 \\
       \aemodel-SD~(no-sup)  &  \textbf{0.904}  & \textbf{0.851} \\
       SD-AE-UNet  &  0.779  & 0.766 \\
       SD-AE-UNet (ft) & 0.670 & 0.681 \\
       % \hline
       % \aemodel-SD~(512) & 0.868 & 
    \end{tabular}
    \caption{IOU between models predicted masks and the two reference sets -- ICON's prediction~(as ICON, \cite{tpami:2022:icon}) and human annotated foreground masks~(as human).  \label{tab:iou}
    }
\end{table}

\begin{table}[h!]
    \centering
    \begin{tabular}{c|c|c|c}
       Dataset & FID & CLIP Score & IOU~(human)  \\
       \hline 
       \dataset  & \textbf{10.51} & \textbf{0.251} & \textbf{0.799} \\
       \dataset~(U) & 13.82 & 0.234  & 0.736
    \end{tabular}
    \caption{Metrics for \aemodel-SD trained on \dataset and a unfiltered variant of \dataset with same size, \dataset~(U) }
    \label{tab:filter}
    \vspace{-0.1in}
\end{table}

%The capability of compositing/synthesizing high resolution images are vital in applying diffusion-based generative models for real-world applications. 
Figure~\ref{fig:samples_512} delves into components $(F, B, m)$ of images generated from \aemodel-SD~(512). Take SD samples in the bottom row into account, we find \aemodel-SD~(512) generates samples with comparable quality as SD-v1-4. In contrast to the static images generated by SD-v1-4, the $F, B, m$ components coming with \aemodel-SD simplify downstream image editing tasks. 

\subsection{Quantitative Results}
The rest of this section presents quantitative comparisons on image quality, mask quality, and text-image relevance.

\label{sec:results}

\vspace{2pt}
\noindent \textbf{Image quality}
Here we demonstrate that \aemodel-SD achieves superior composition quality compared to baseline methods. 
% Following previous works on image generation, we take the key automated metrics FID~\cite{nips:2017:fid} to measure image fidelity. We report FID scores on two test sets. The major one is a 20000 examples held-out test subset of \dataset. Plus, we also quantify models ability to generate diverse images beyond \dataset with a 20000 unfiltered subset LAION-Aesthetic. This subset does not overlap with \dataset. We abbreviate the two datasets as LL\textsuperscript{2}I and LA respectively. 
We present in Table~\ref{tab:fid_clip}~(in the 4th and 5th column) FID scores of models on both test sets. Our first observation is that \aemodel-SD outperforms baseline models in generating $256 \times 256$ two-layer images by big margins, which are around 3.5 for LL\textsuperscript{2}I set and 5 for LA set. We attribute it to the distributional misalignment between the $(F, B, m)$ components of a two-layer image and a natural image used in SD. Such distributional misalignment could cause larger errors in reconstructing $F$, $B$, and $m$ from the latent space of SD's autoencoder. In contrast, \aemodel's decoder decompresses $F$, $B$, and $m$ with separate prediction branches, making better usage of the decoder's capacity.

Meanwhile, \aemodel-SD achieves a lower FID compared to SD-\dataset on both test sets, which indicates that the composed image quality benefits from the layered-image generation training compared to the text2image setting. It is possible that the mask in the layered-image brought additional information that benefits the composed image quality. \aemodel-SD (512) achieves better results than \aemodel-SD, which sets the promise that training on a higher resolution could further improve the image quality significantly. We also notice a considerable improvement from \aemodel-SD~(no-sup) to \aemodel-SD. This validates our hypothesis that the supervision of the original image $I$ can enhance generative quality.
% That improvement verifies our hypothesis that the full image partially reliefs the noisy estimations of $(F, B, m)$ due to our imperfect dataset creation.

Lastly, we showed the results of SD-v1-4~\cite{cvpr:2022:ldm}, which was trained on a much larger dataset, for reference. The FID of SD-v1-4 is not directly comparable with other approaches in the table, but \aemodel-SD still achieves lower FID compared to the SD on LL\textsuperscript{2}I set for the resolution of $512$, demonstrating the success of our whole two-layer image generation pipeline. Though in the fifth column, we find \aemodel-SD has worse FID compared to the SD-v1-4 on a more broad LA set, we believe this is reasonable considering our current \dataset is less than $1/10$ as the size of LAION-Aesthetics and less than $1/30$ as the size of LAION-2B. For future work, we will create a larger version of \dataset to eliminate the gap. 

%\vspace{2pt}
%\noindent \textbf{Analysis} \todo{need Xin tell an exciting story.} We believe the exciting result of \aemodel-SD is achieved via our careful model design. We present a brief analysis here. One the one hand, we see from Table~\ref{tab:image_quality} that crafting an autoencoder specialized for two-layer image generation is better than na\"ively leveraging Stable Diffusion's original architecture. We attribute it to the distributional misalignment between the $(F, B, m)$ components of a two-layer image and a natural image used in SD. Such distributional misalignment could cause larger errors in reconstructing $F$, $B$, and $m$ from the latent space of SD's autoencoder. In contrast, \aemodel's decoder decompresses $F$, $B$, and $m$ with separate prediction branches, making better usage of the decoder's capacity. On the other hand, 

% Table~\ref{} indicates that \todo{ ... } \todo{compare with SD}
% Besides, since XXX is targeting at composition-based generation, we expect its has slightly lower generation quality compared to the general text-to-iamge generative models. 
% Therefore, we also include comparisons with open-source image-to-text generative models Stable Diffusion~\cite{cvpr:2022:ldm}, YYY, and ZZZ to show the gap. 

\vspace{2pt}
\noindent \textbf{Mask quality}
% The quality of mask $m$ is the central metric in image matting and plays a crucial role in high-quality image composition. We want to evaluate it on \aemodel-SD and baselines. Yet, the alpha mask in \dataset is estimated without ground-truth, we will compare the their mask predictions with the prediction ICON to measure how much the alpha channel capture the important regions in the image. Furthermore, we also compare generative models' mask predictions with a state-of-art referred segmentation model \todo{name} . 
We examine the quality of the masks produced by two-layer image generative models. The IOUs highlighted in Table~\ref{tab:iou} show that compared to other baselines, \aemodel-SD and \aemodel-SD~(no-sup) produce masks that are more accurate according to the reference ICON~\cite{tpami:2022:icon} salient detection model. In addition, their superior IOU on the human-annotated test set shows \aemodel-SD's masks are better at capturing foreground semantics compared to other baselines. \aemodel-SD shows a slightly worse number than \aemodel-SD~(no-sup) indicating that the usage of the composed image in the autoencoder training drives better image quality but might hurt mask quality. That observation introduces future work to better balance each layer's generation.
%\todo{We noticed that \aemodel-SD slightly worse than \aemodel-SD~(no-sup) on this evaluation. However, since their margins with other methods are large, we are fine with it. }
 %\todo{some descriptions ...}
% For the last, We use Adobe Photoshop to produce a coarse ground-truth and measure their differences with models' prediction. We report SAD in Table~\ref{}. 

\vspace{2pt}
\noindent \textbf{Image-text relevance} As depicted in Table~\ref{tab:fid_clip}~(in the last column), the proposed \aemodel-SD outperforms all baseline models in the $256 \time 256$ resolution. \aemodel-SD (512) achieves a better text-image relevance compared to \aemodel-SD, which indicates a promising performance improvement when training on a higher resolution. Similarly, we list the SD-v1-4's results for reference, and the proposed \aemodel-SD's result is on par with SD-v1-4's results even if the SD-v1-4 was trained on a much larger dataset. 
%For the resolution of 512, \aemodel-SD's score is just slightly below that of the SD model, which is reasonable as already discussed discrepancy between the size of proposed \dataset and LAION-Aesthetics.
% \footnote{\url{https://laion.ai/blog/laion-aesthetics/}} 
%\todo{link in the footnote, less than one-tenth? response: yes, and we've footnoted in dataset section} 
The results suggest that the combination of our efforts on the dataset and model architecture ensures the generated two-layer images follow the instruction of text prompts. 

% Finally, we validate that images composited by our \aemodel-SD faithfully reflect text prompts. Following previous works~(e.g.\cite{cvpr:2022:ldm}), we use CLIP~\cite{icml:2021:clip} score as a surrogate metric for image-text relevance. We calculate CLIP score with the 20000 held-out subset of \dataset. 

\subsection{Effectiveness of Data Filtering}
\label{sec:eval_filter}

Large-scale high-quality datasets are indispensable to the success of deep neural networks in computer vision. Besides the quality metrics shown above, here we demonstrate the effect of our data filtering strategies proposed in Section~\ref{sec:quality_filter}. Illustrated in Table~\ref{tab:filter}, the large improvement of FID, CLIP score, and IOU for \aemodel-SD on the \dataset and its unfiltered variant \dataset~(U) demonstrates the effectiveness of the proposed dataset synthesis approach discussed in Section~\ref{sec:dataset}.

% \subsection{Ablation Study}

%\subsection{Discussion}

\section{Conclusions and Discussions}
In this paper, we proposed a layered-image generation problem and scoped it to a two-layer image generation setting. We created a $57.02M$ high-quality layered-image dataset and used it to build a layered-image generation system using latent diffusion models. We designed the evaluation metric for the layered-image generation and demonstrated a benchmark in terms of image quality, image-text relevance, and mask quality.

Additionally, the proposed method is not limited to two layers, which can be applied to any fixed number of layers. Meanwhile, a conditional model can be developed as a natural extension of this work, which potentially could generate a layer given existing layers and that can further enable layered image generation of an arbitrary number of layers. We leave it as a future work.

{\small
\bibliographystyle{ieee_fullname}
\bibliography{main}
}

\clearpage
\begin{appendices}

\section*{\Large{Appendix I: More Results}}

In Section 5.4, Figures 5 and 6 illustrate the powerful generative ability of \aemodel-SD. We visualize more samples in this section. 

\begin{figure}[h]
    \centering
    \includegraphics[width=\linewidth]{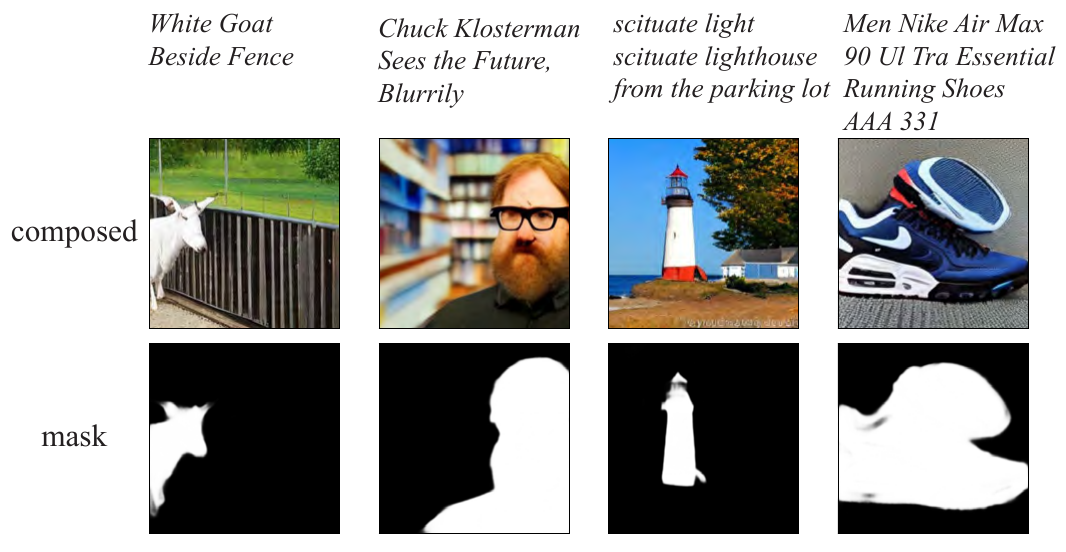}
    \caption{More 256 samples for \aemodel-SD on \dataset test set. Prompts are displayed at the top. }
    \label{fig:256_more_laion_1}
\end{figure}

\begin{figure}[h]
    \centering
    \includegraphics[width=\linewidth]{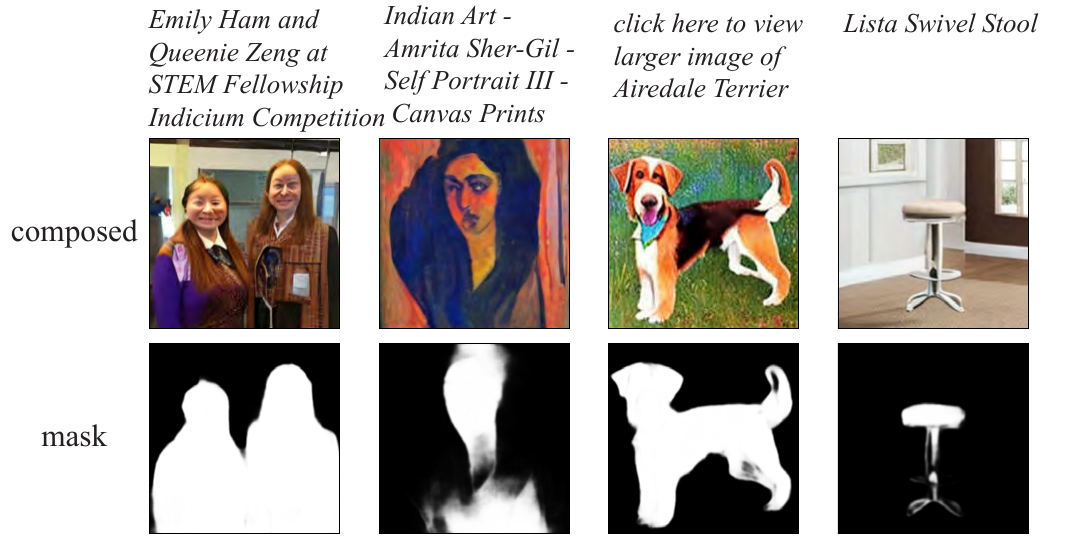}
    \caption{Some failed generations from \aemodel-SD. Prompts are displayed at the top. }
    \label{fig:failure_cases}
\end{figure}

\begin{figure}
    \centering
    \begin{subfigure}{\linewidth}
    \includegraphics[width=\linewidth]{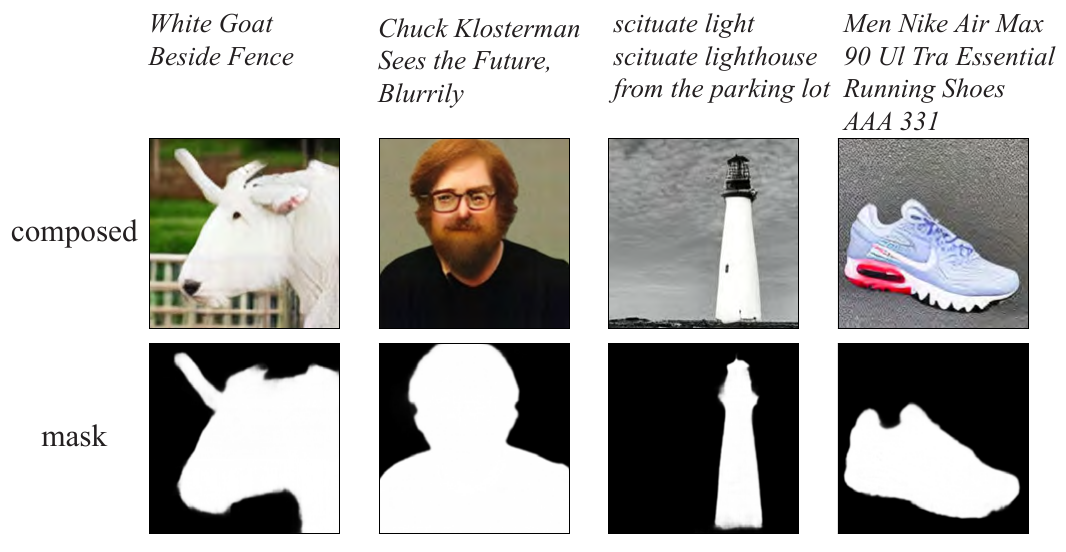}
    \caption{\aemodel-SD (no-sup)}
    \end{subfigure}
    \begin{subfigure}{\linewidth}
    \includegraphics[width=\linewidth]{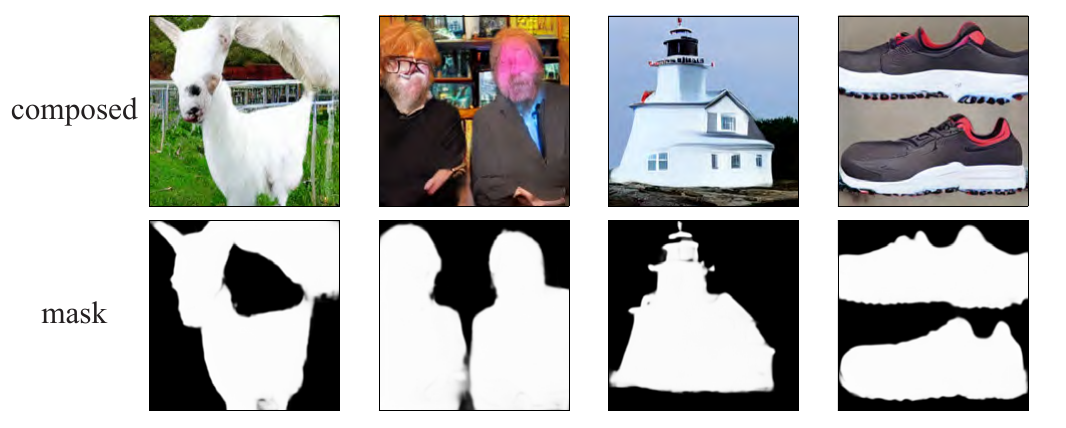}
    \caption{SD-AE-UNet}
    \end{subfigure}
    \begin{subfigure}{\linewidth}
    \includegraphics[width=\linewidth]{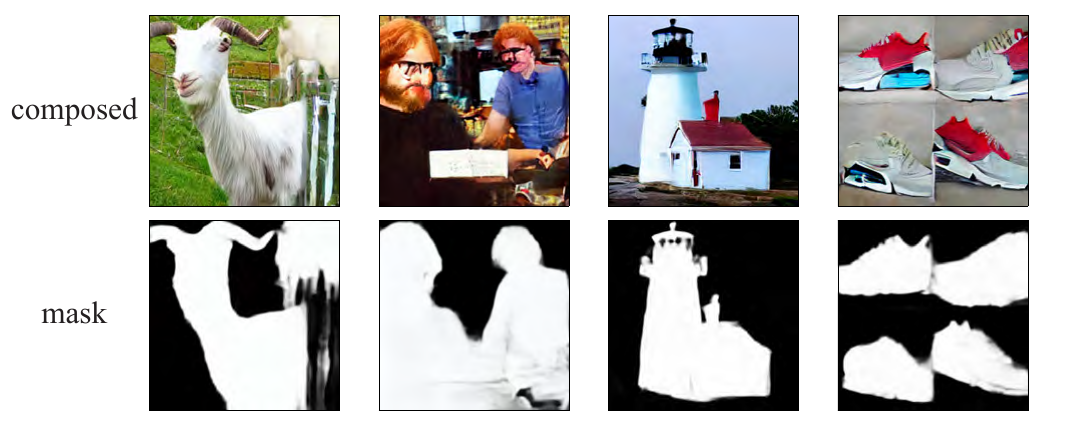}
    \caption{SD-AE-UNet (ft)}
    \end{subfigure}
    \caption{More 256 samples for baseline methods on \dataset test set. Prompts are displayed at the top. }
    \label{fig:256_more_laion_2}
\end{figure}

\begin{figure*}
    \centering
    \includegraphics[width=0.9\linewidth]{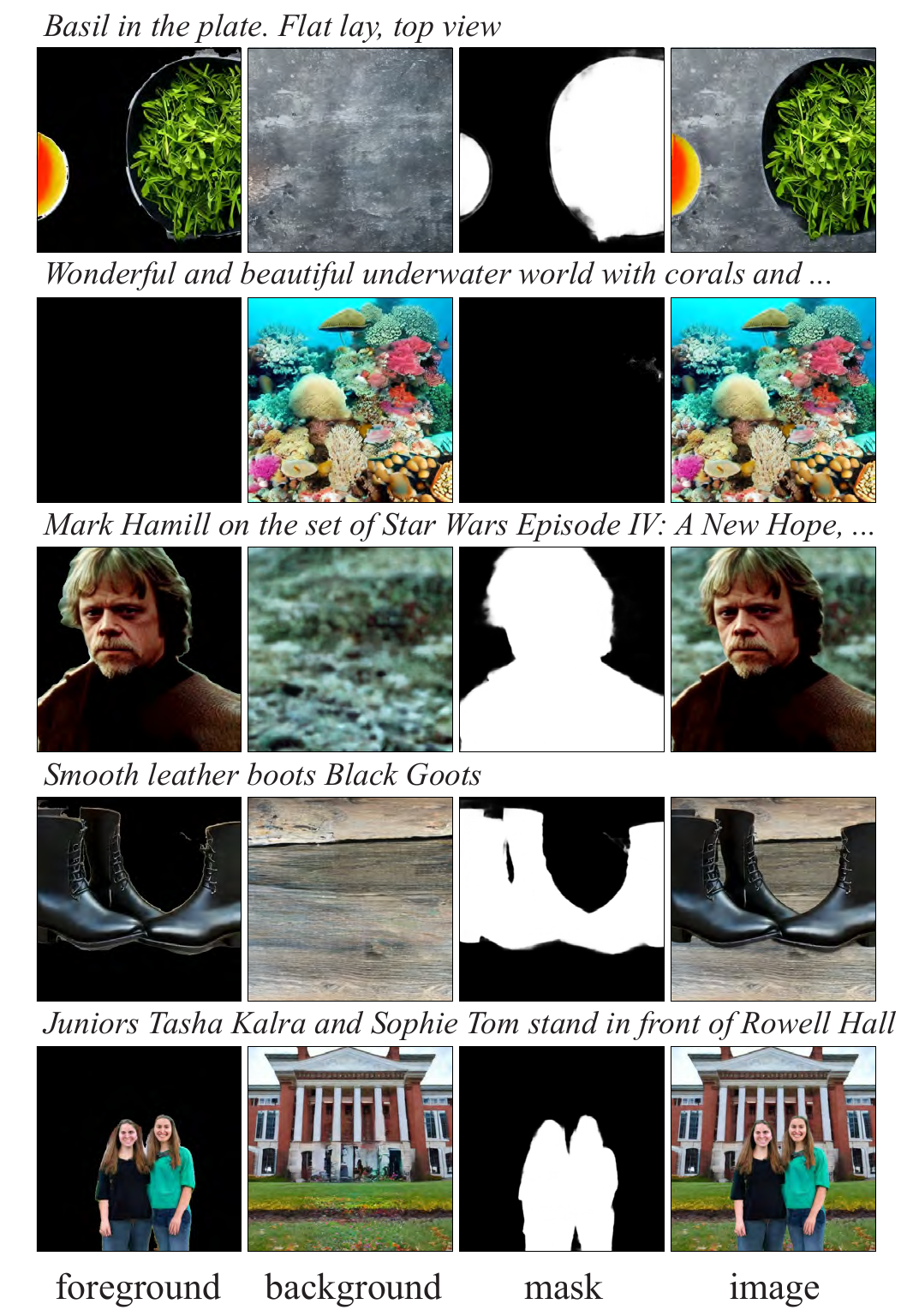}
    \caption{More \aemodel-SD~(512) synthesized two-layer images. The prompts above images are from \dataset test set. }
    \label{fig:512_more_laion}
\end{figure*}

\begin{figure*}
    \centering
    \includegraphics[width=0.9\linewidth]{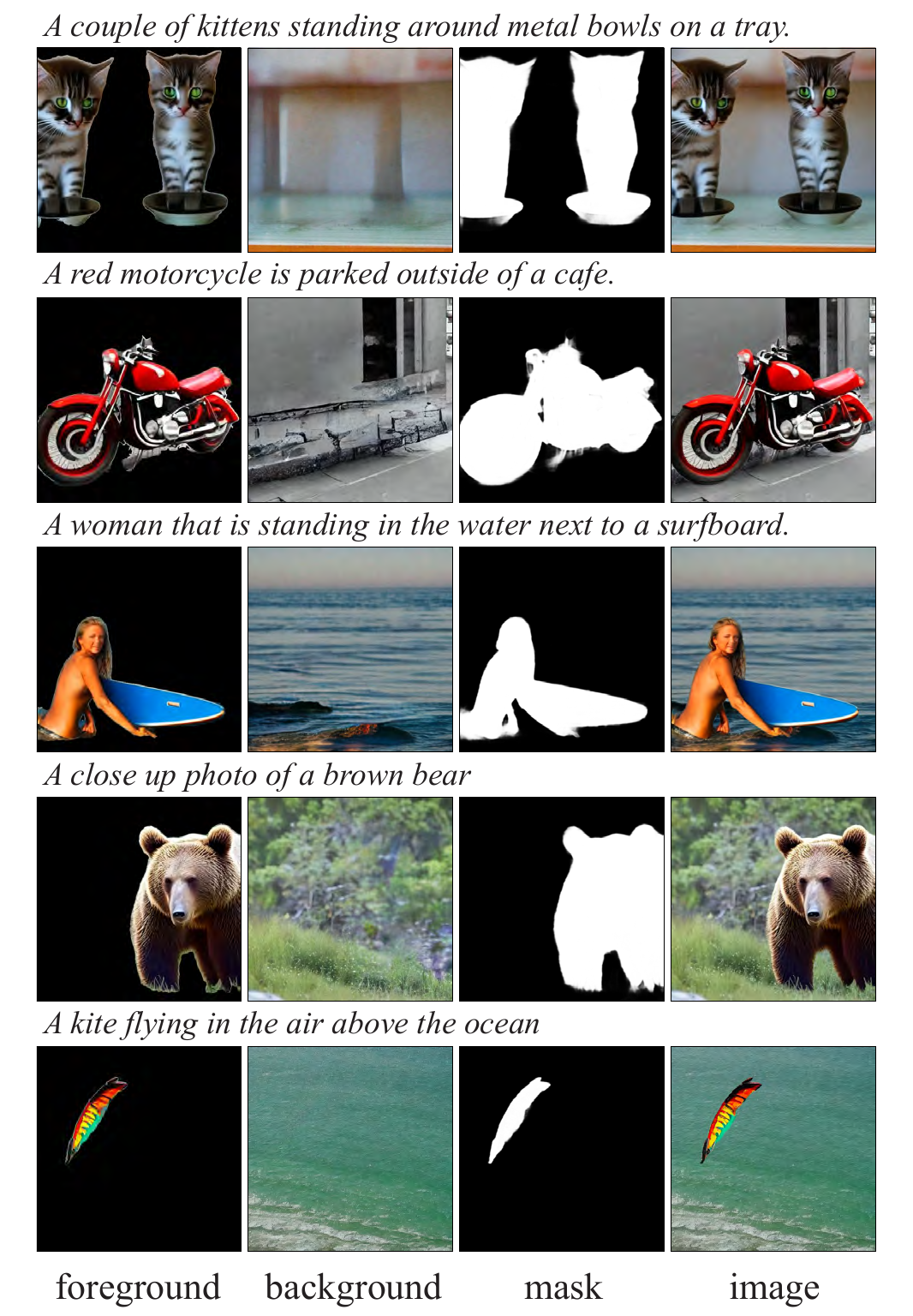}
    \caption{More \aemodel-SD~(512) synthesized two-layer images. The prompts above images are from MSCOCO 2014~\cite{eccv:2014:mscoco} validation set. }
    \label{fig:512_more_mscoco}
\end{figure*}

\subsection{More 256 samples}
We show more $256 \times 256$ two-layer images of our \aemodel-SD and baseline methods on \dataset test set in Figures~\ref{fig:256_more_laion_1} and \ref{fig:256_more_laion_2}.

\subsection{\aemodel-SD~(512) samples}

Figure~\ref{fig:512_more_laion} provides samples of \aemodel-SD~(512) with prompts from test set of \dataset. In addition, Figure~\ref{fig:512_more_mscoco} displays samples of \aemodel-SD~(512) with prompts from more commonly used MSCOCO dataset~\cite{eccv:2014:mscoco}. 

\subsection{Failure cases}

We present some failure cases of \aemodel-SD in Figure~\ref{fig:failure_cases}. We noticed that \aemodel-SD sometimes predicts objects with the wrong geometry~(faces in the first column) or structures~(the dog in the third column). Besides, some masks~(the second and fourth columns) are inaccurate.  

\section*{\Large{Appendix II: Dataset}}

\begin{figure}
    \centering
    \includegraphics[width=\linewidth]{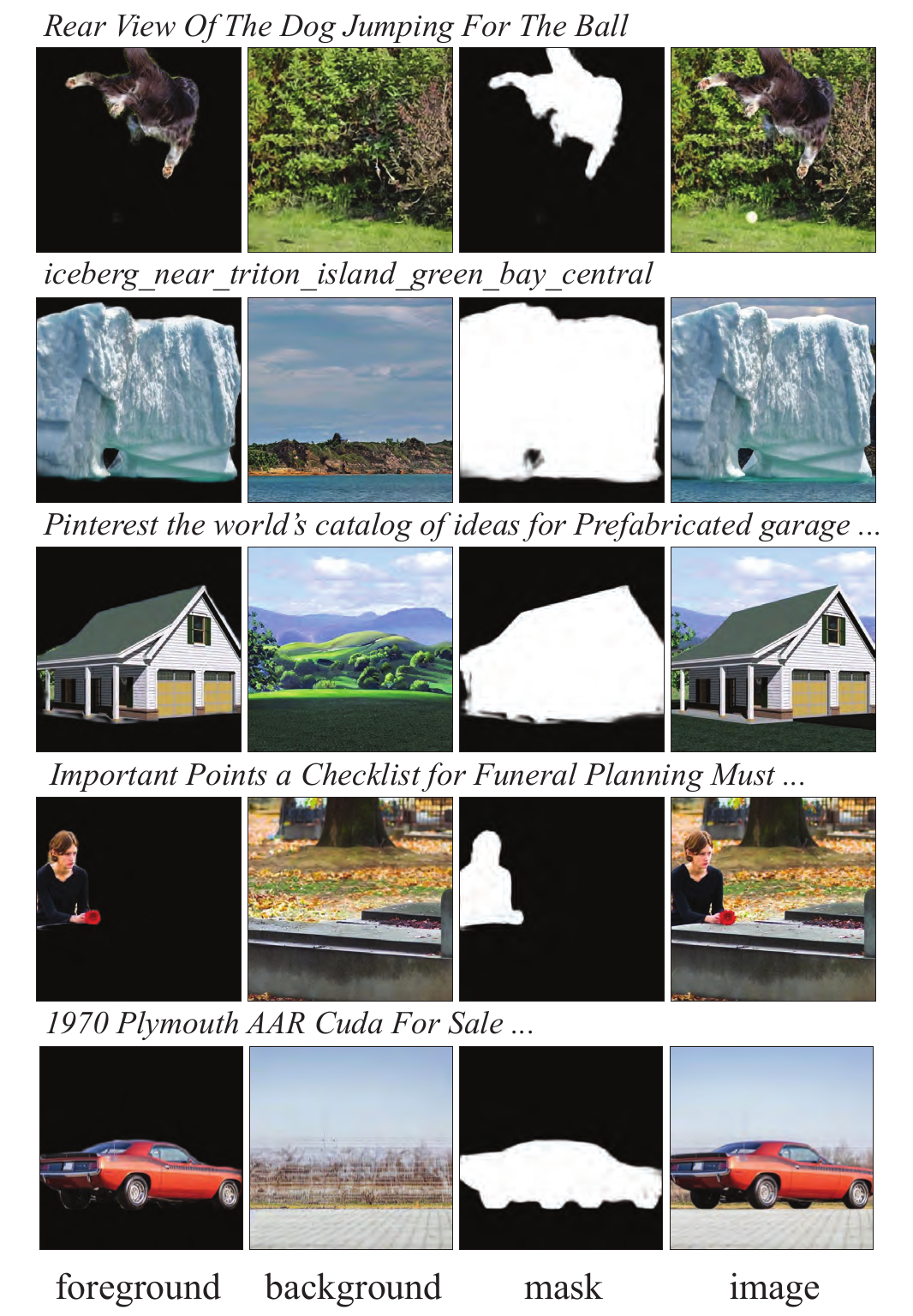}
    \caption{Samples of two-layer images from our \dataset dataset. Prompts are on the top of images.  }
    \label{fig:ours_more_examples_pg1}
\end{figure}

\begin{figure}
    \centering
    \includegraphics[width=\linewidth]{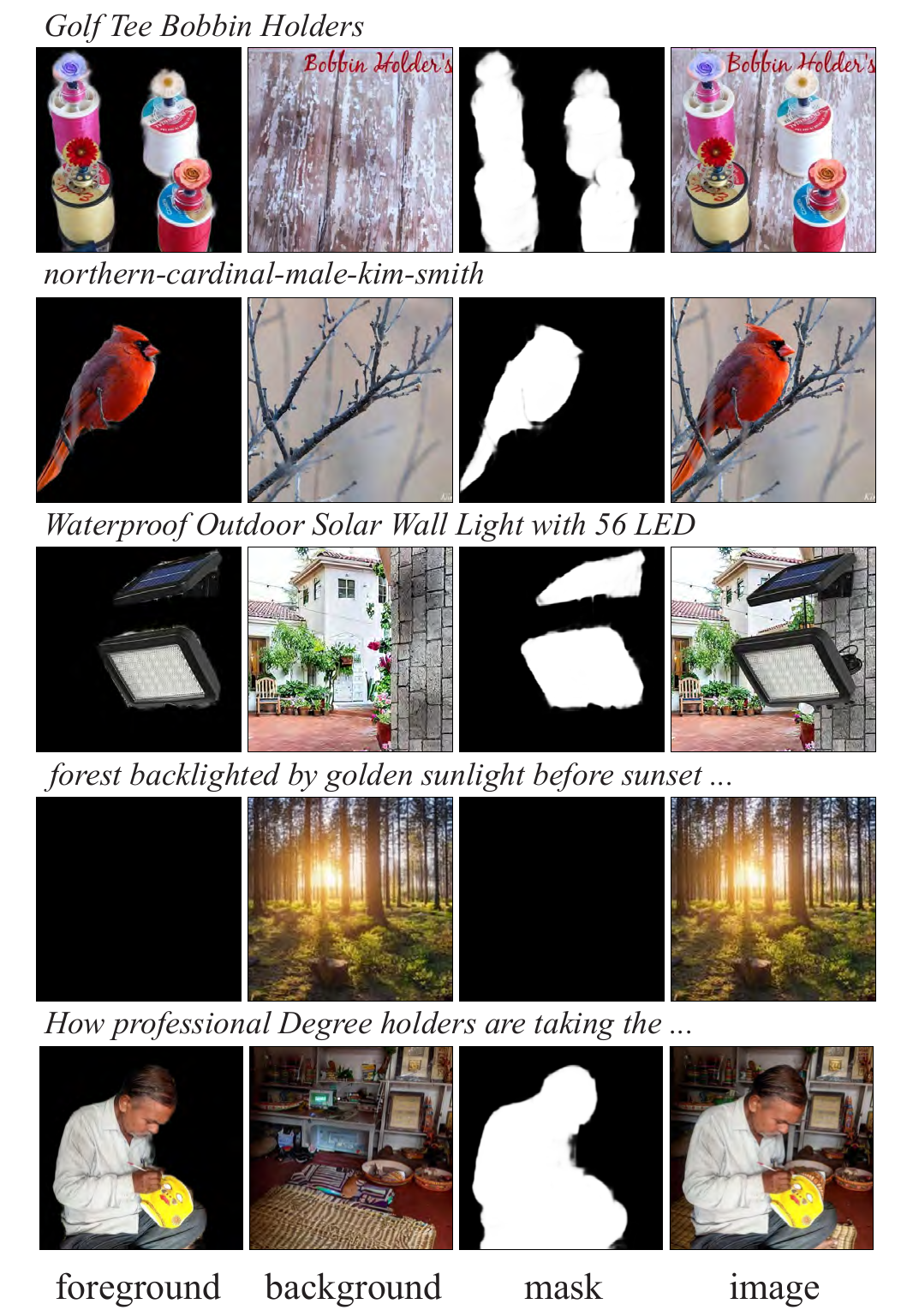}
    \caption{More samples of two-layer images from our \dataset dataset. Prompts are on the top of images.  }
    \label{fig:ours_more_examples_pg2}
\end{figure}

\begin{figure*}[t]
    \centering
    \begin{subfigure}{\linewidth}
        \includegraphics[width=\linewidth]{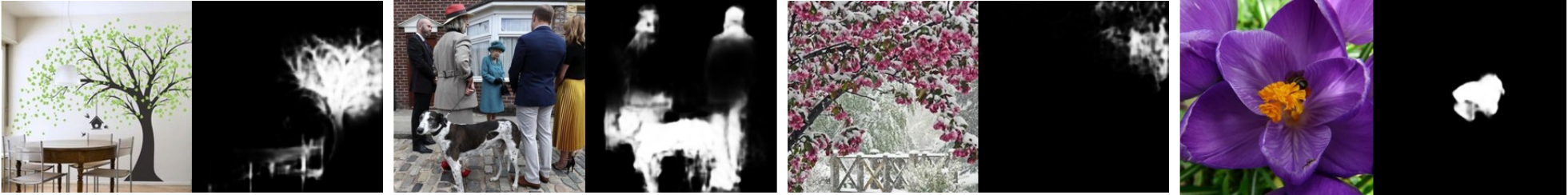}
        \caption{Predicted ``bad'' masks}
    \label{fig:salient_quality_before}
    \end{subfigure}
    \begin{subfigure}{\linewidth}
        \includegraphics[width=\linewidth]{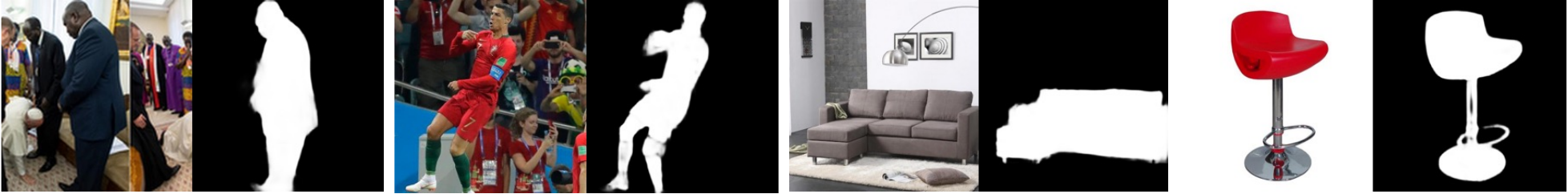}
        \caption{Predicted ``good'' masks}
    \label{fig:salient_quality_after}
    \end{subfigure}
    \begin{subfigure}{\linewidth}
        \includegraphics[width=\linewidth]{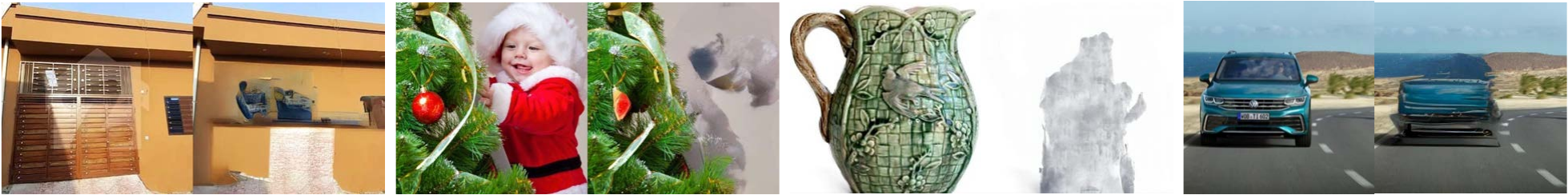}
        \caption{Predicted ``bad'' inpaintings}
    \label{fig:inpainting_quality_before}
    \end{subfigure}
    \begin{subfigure}{\linewidth}
        \includegraphics[width=\linewidth]{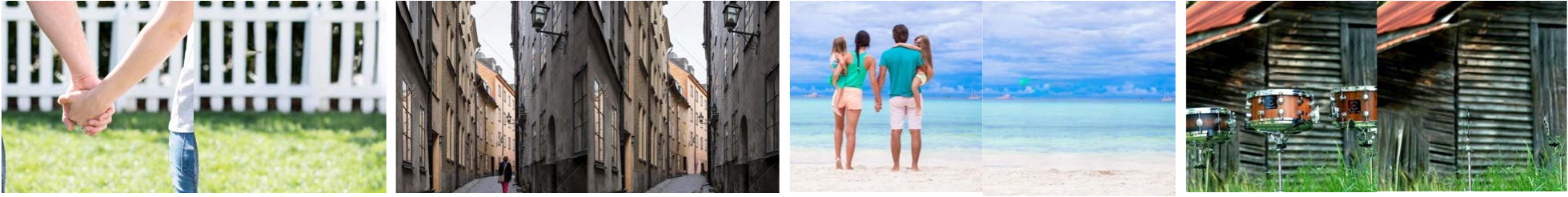}
        \caption{Predicted ``good'' inpaintings}
    \label{fig:inpainting_quality_after}
    \end{subfigure}
    
    \caption{Predicted good and bad salient masks and inpaintings}
    \label{fig:salient_quality}
% \vspace{-0.2in}
\end{figure*}

\subsection{Data filtering}
In Section 3.4, we discussed the filtering of low-quality masks and inpaintings to ensure the final quality of our \dataset dataset. More examples of before and after data filtering are visualized in Figure~\ref{fig:salient_quality}.

\subsection{More examples}

Figures~\ref{fig:ours_more_examples_pg1} and \ref{fig:ours_more_examples_pg2} depict ten more two-layer images from our $57.02M$ \dataset dataset.

\subsection{Analysis and statistics}

As a complement to Section 3, we present several quantitative analyses of \dataset and LAION-Aesthetics~(short for LAION-A\footnote{\url{https://laion.ai/blog/laion-aesthetics/}}~\cite{nips:2022:laion5b}) here. The analysis focuses on the complexity of images, masks, and text prompts. Plus,we also compare the text-image relevance of \dataset and LAION-A. 

Since the magnitude of both datasets are large, we sample $100K$ examples randomly from each dataset for following analyses.  

\begin{figure}
    \centering
    \begin{subfigure}{\linewidth}
    \includegraphics[width=\linewidth]{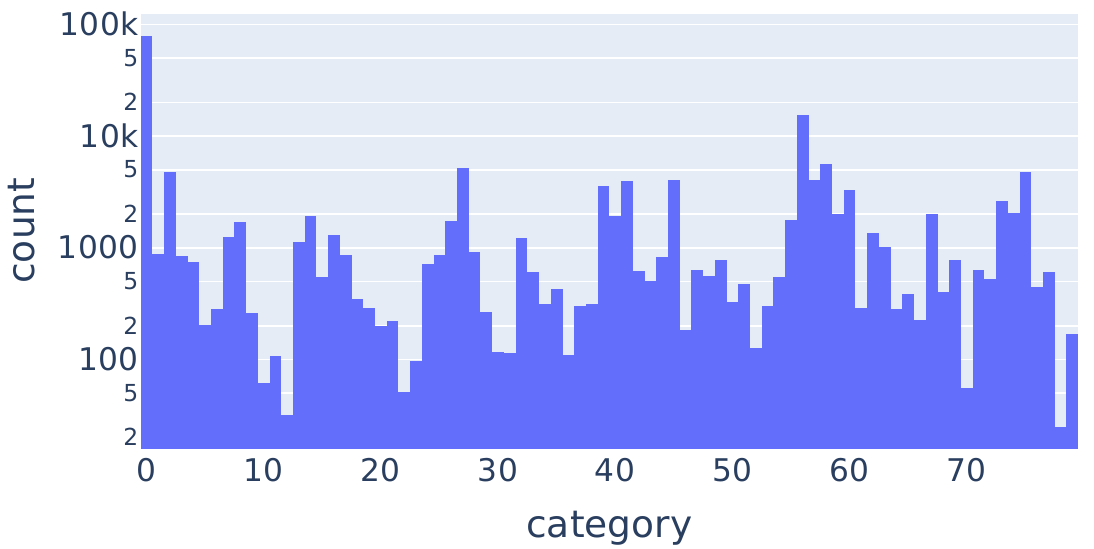}
    \caption{LAION-A \label{fig:obj-per-cat-laion-a}}
    \end{subfigure}
    \begin{subfigure}{\linewidth}
    \includegraphics[width=\linewidth]{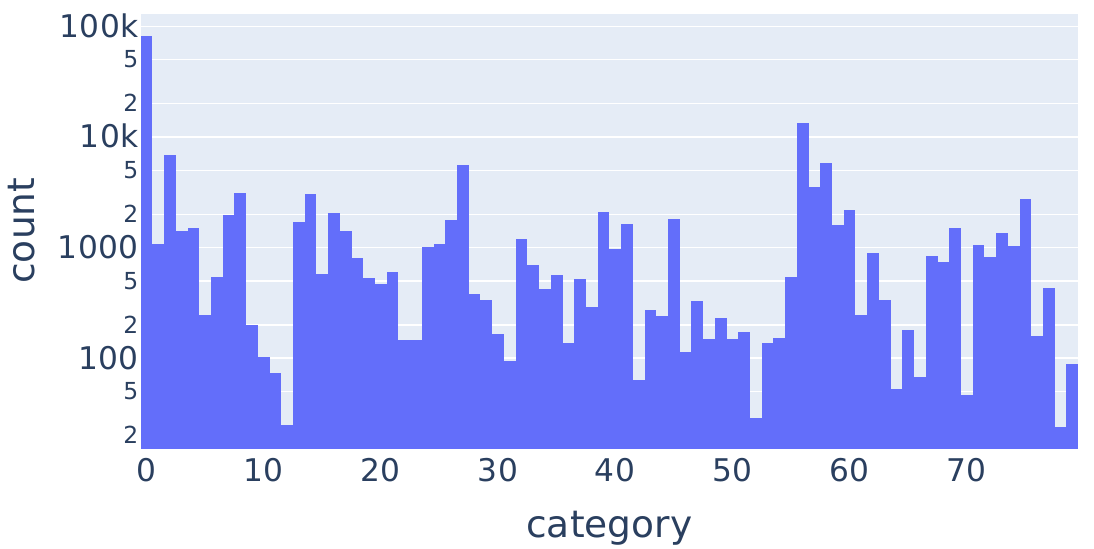}
    \caption{\dataset \label{fig:obj-per-cat-ours}}
    \end{subfigure}
    \caption{Number of objects per MSCOCO 80 category in LAION-A~\subref{fig:obj-per-img-laion-a} and \dataset~\subref{fig:obj-per-img-ours}. 
    \label{fig:obj-per-cat}}
\end{figure}

\begin{figure}
    \centering
    \begin{subfigure}{0.45\linewidth}
    \includegraphics[width=\linewidth]{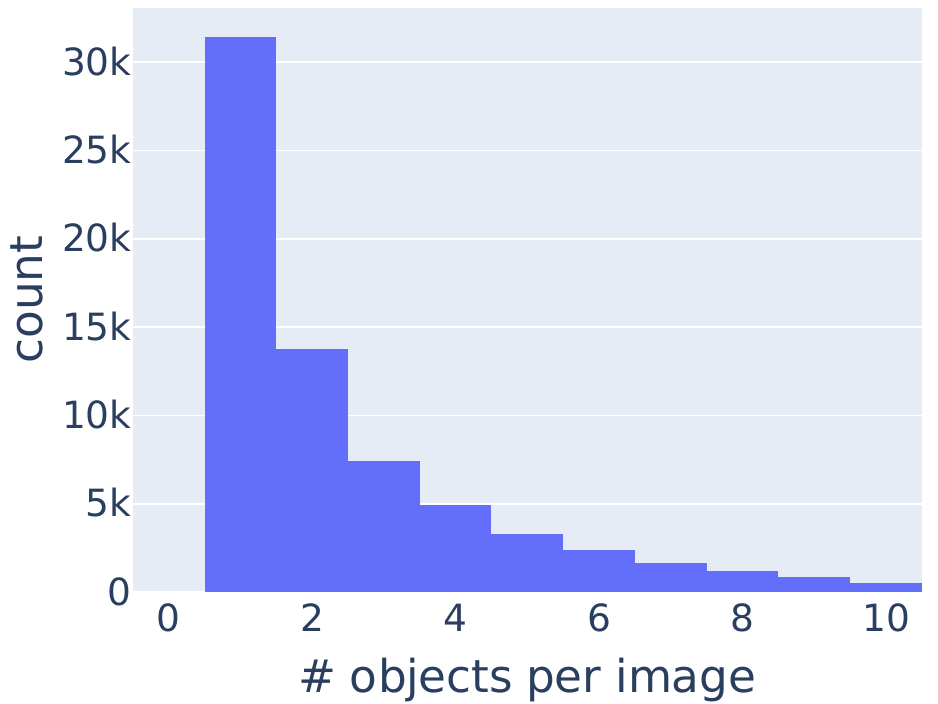}
    \caption{LAION-A \label{fig:obj-per-img-laion-a}}
    \end{subfigure}
    \begin{subfigure}{0.45\linewidth}
    \includegraphics[width=\linewidth]{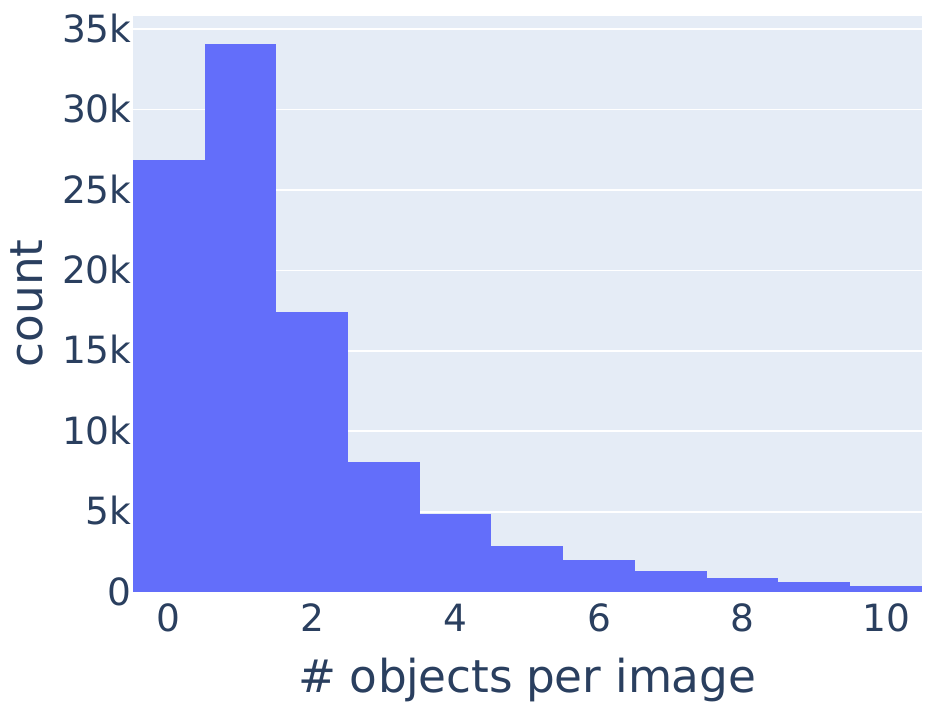}
    \caption{\dataset \label{fig:obj-per-img-ours}}
    \end{subfigure}
    \caption{ Number of objects per image in LAION-A~\subref{fig:obj-per-img-laion-a} and \dataset~\subref{fig:obj-per-img-ours}. 
    \label{fig:obj-per-img}}
\end{figure}

\vspace{2pt}
\textbf{Objects}. We apply Yolov7~\cite{arxiv:2022:yolov7}, a high-performance object detector trained on MSCOCO~\cite{eccv:2014:mscoco} dataset, on both datasets and obtain object categories statistics~(in Figure~\ref{fig:obj-per-cat}) and the number of objects per image statistics~(in Figure~\ref{fig:obj-per-img}). Though MSCOCO has limited object categories and fewer diverse images compared to LAION-5B. We argue that the above analysis with Yolov7 partially reflects that images retained in \dataset have similar complexity to those of LAION-A in terms of ``objects.'' 

We observe that both distributions of number of objects per category and number of objects per image on our \dataset dataset and MSCOCO are similar. For Figure~\ref{fig:obj-per-cat}, only a few categories have a noticeable difference in terms of the number of objects. For Figure~\ref{fig:obj-per-img}, \dataset has slightly more images with two objects compared with LAION-A which has more images of zero MSCOCO objects.

\begin{figure}
    \centering
    \begin{subfigure}{\linewidth}
    \includegraphics[width=\linewidth]{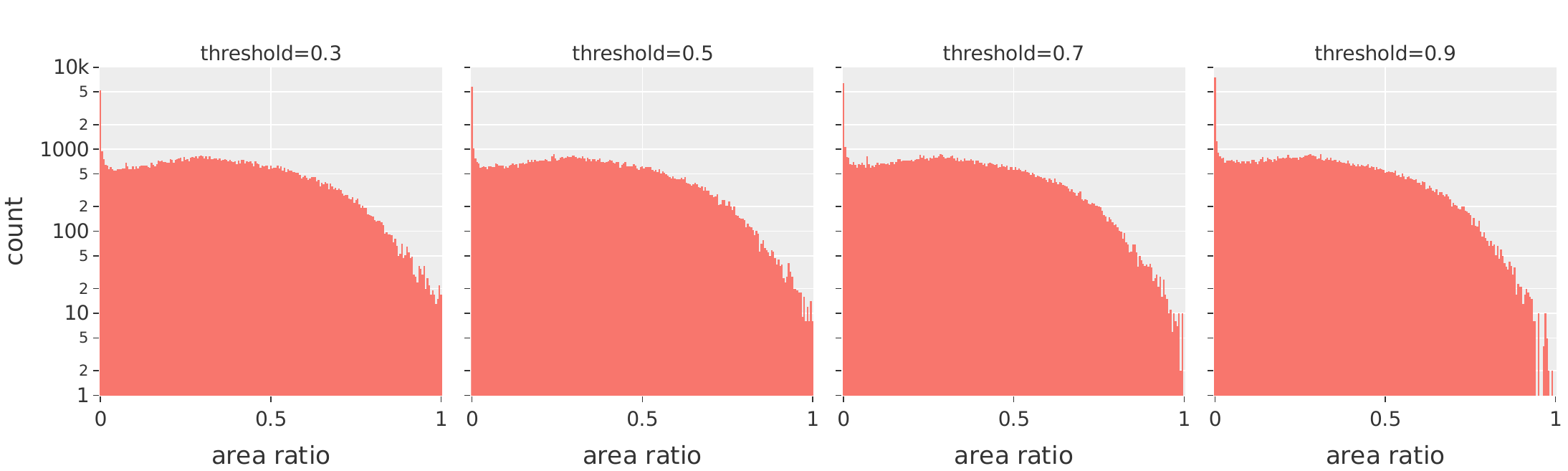}
    \caption{LAION-A}
    \label{fig:area-ration-laion-a}
    \end{subfigure}
    \begin{subfigure}{\linewidth}
    \includegraphics[width=\linewidth]{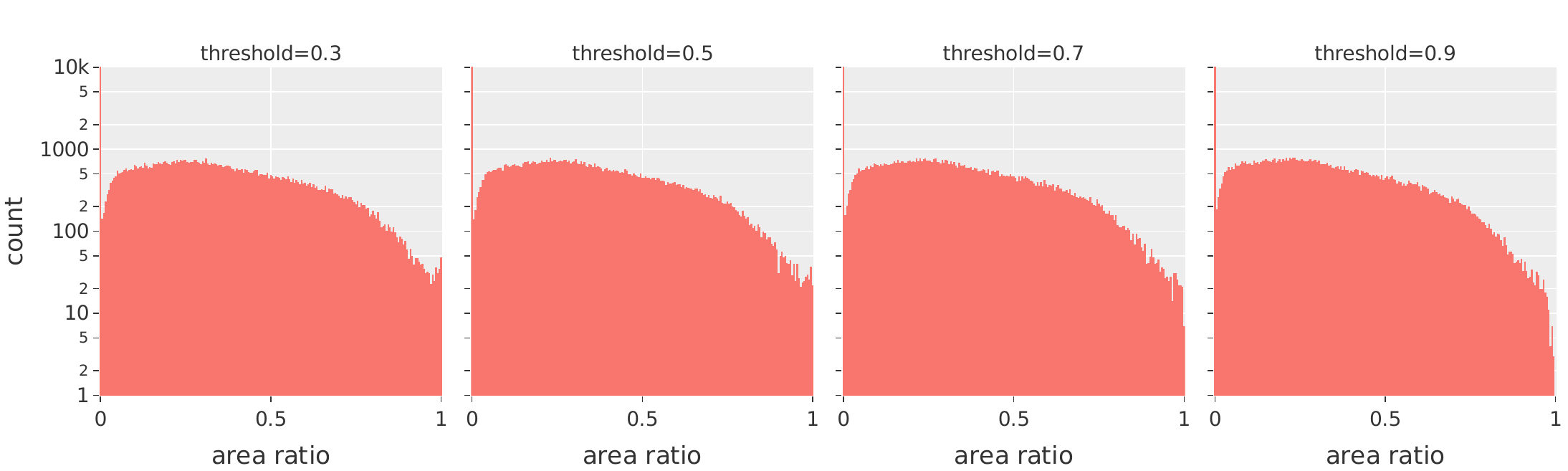}
    \caption{\dataset}
    \label{fig:area-aratio-ours}
    \end{subfigure}
    \caption{Histograms of area of masks (normalized to $[0, 1]$) predicted by ICON. We take four different thresholds for binarizing masks. }
    \label{fig:area-ratio}
\end{figure}

\begin{figure}
    \centering
    \begin{subfigure}{\linewidth}
    \includegraphics[width=0.95\linewidth]{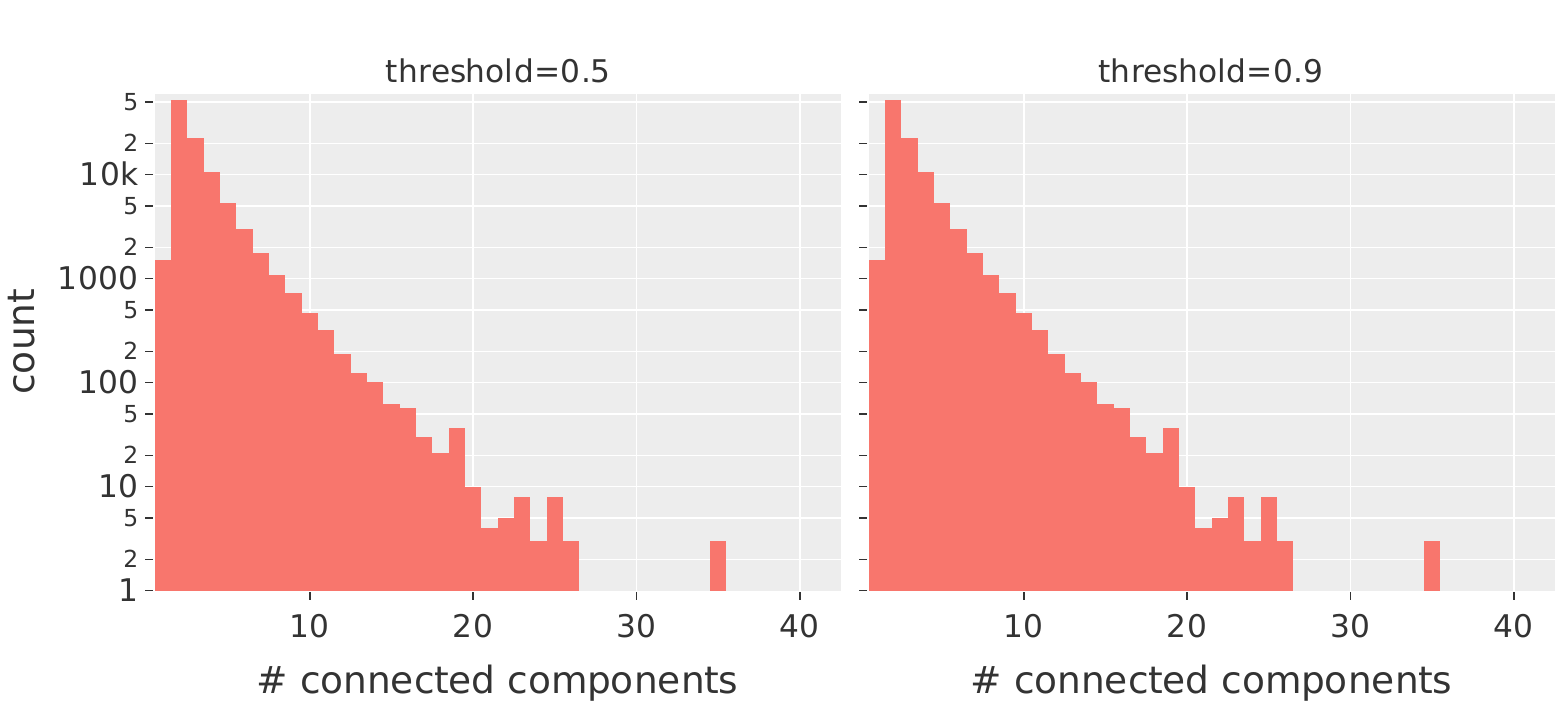}
    \caption{LAION-A}
    \label{fig:ncc-laion-a}
    \end{subfigure}
    \begin{subfigure}{\linewidth}
    \includegraphics[width=0.95\linewidth]{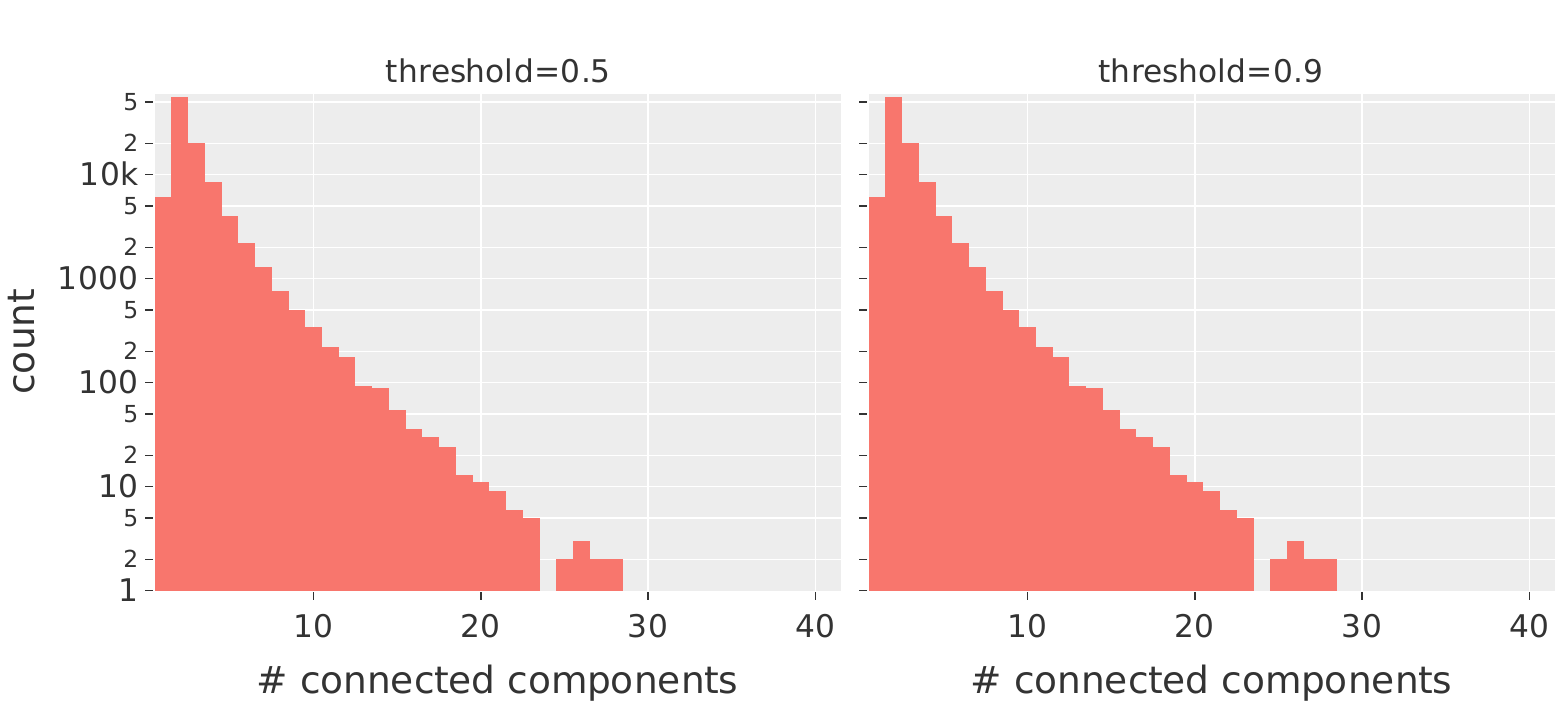}
    \caption{\dataset}
    \label{fig:ncc-ours}
    \end{subfigure}
    \caption{Histograms of number of 8-way connected components in masks predicted by ICON. We take two different thresholds for binarizing masks. We filter out connected components whose areas are smaller than 100. }
    \label{fig:ncc}
\end{figure}

% \vspace{-2pt}
\textbf{Mask}. In constructing \dataset, we leverage ICON~\cite{tpami:2022:icon} to estimate masks $m$ from images in LAION-A. Again, we compare statistical differences between \dataset and LAION-A. We first study the area ratio of salient regions in images under various binary thresholds. As illustrated in Figure~\ref{fig:area-ratio}, the distributions of the area ratio of masks are similar for \dataset and LAION-A across all the thresholds. Our second study concerns the number of connected components for masks, which indicates scattering patterns of objects and the noise level of masks. Still, as shown in Figure~\ref{fig:ncc}, the distributions of the two datasets are almost the same.

\begin{table}
    \centering
    \begin{tabular}{c|c|c}
      percentile ($p\%$)& length~(LAION-A) & length~(\dataset) \\
      \hline
      1  & 1  & 1 \\
      10  & 3 & 3 \\
      25 & 5  & 5 \\
      50 & 7 &  7\\
      75 & 10 & 10 \\
      90 & 15 & 15\\
      99 & 36 & 36
    \end{tabular}
    \caption{Percentiles of the number of tokens in text prompts of LAION-A and dataset. Stopwords, symbols, and punctuation are removed for counting.  }
    \label{tab:caption_length}
\end{table}

\begin{table}
    \centering
    \begin{tabular}{c|c}
       dataset  & CLIP score \\
       \hline
        LAION-A & $0.278 \pm 0.037 $ \\
        \dataset & $0.273 \pm 0.034 $ 
    \end{tabular}
    \caption{CLIP scores for pairs of image and prompts from LAION-A and \dataset. }
    \label{tab:clip}
\end{table}

\vspace{2pt}
\textbf{Prompt}. For the text prompts, we compare the lengths and the word frequency on \dataset and LAION-A. Table~\ref{tab:caption_length} displays caption lengths at different percentiles. It is clear that both datasets have prompts with similar lengths. Furthermore, to gain a word-level understanding, we calculate the Pearson correlation coefficient for frequencies of the top $5K$ words between these two datasets, the value is $0.930$, suggesting \dataset roughly represents a uniformly distributed subset of LAION-A. Finally, we quantify the image-text relevance with CLIP score~\cite{icml:2021:clip} as we have done in Section 4. We notice from Table~\ref{tab:clip} that the two datasets have similar CLIP scores.

\section*{\Large{Appendix III: Implementation Details}}

We detail the loss terms used in Section~4.3 for training our \aemodel. 

\vspace{2pt}
\textbf{Image loss}. For the supervision of image components $(F, B, I)$ , we use the same loss terms as used in Latent Diffusion~\cite{cvpr:2022:ldm}. The loss terms are separately applied to each of $F, B, I$ with equal weights.  

\vspace{2pt}
Inspired by literature of image 
 matting~\cite{cvpr:2017:deepmat,eccv:2021:mgm,aaai:2020:gca-matting,cvpr:2019:ctx-mat}. We enforce composition loss~\cite{cvpr:2017:deepmat} and Laplacian loss~\cite{cvpr:2019:ctx-mat} and for the \aemodel decoder's predicted mask $\hat{m}$ and the ground-truth mask $m$.

\vspace{2pt}
\textbf{Composition loss}. Let $C$ be the composed image of $(F, B, m)$, in other word, $C = mF + (1 - m)B$. Then the composition loss~\cite{cvpr:2017:deepmat} is defined as 
\begin{equation}
    \ell_\mathrm{comp} = \sum_{i} \sqrt{(C_i - \hat{C}_i)^2 + \epsilon^2}
\end{equation}
where $i$ run through all the pixels, $\hat{C} = \hat{m}F + (1 - \hat{m}B)$ is the composed image with decoder's predicted mask $\hat{m}$, and $\epsilon$ set to $10^{-6}$ for numerical stability. 

\vspace{2pt}
\textbf{Laplacian loss}. Laplacian loss~\cite{cvpr:2019:ctx-mat} can be used to measure the difference between the predicted mask $\hat{m}$ with the ground-truth mask $m$. Specifically,  
\begin{equation}
    \ell_\mathrm{lap} = \sum_{j=1}^3 2^{j - 1} \| \phi^{j}(\hat{m}) - \phi^{j}(m) \|_1
\end{equation}
where $\phi^{j}(m)$ denotes the $j$-th level of Laplacian pyramid of the mask.

\end{appendices}

\end{document}